\documentclass{article}

\usepackage{microtype}
\usepackage{graphicx}
\usepackage{booktabs} 
\usepackage{mathtools}
\usepackage{array}

\usepackage{hyperref}

\usepackage[accepted]{icml2023}

\usepackage{amsmath}
\usepackage{amssymb}
\usepackage{mathtools}
\usepackage{amsthm}

\usepackage[capitalize,noabbrev]{cleveref}

\theoremstyle{plain}
\newtheorem{theorem}{Theorem}[section]

\newtheorem{lemma}[theorem]{Lemma}

\theoremstyle{definition}

\theoremstyle{remark}

\newcommand{\R}{\mathbb{R}}
\newcommand{\Z}{\mathbb{Z}}
\newcommand{\T}{\mathbb{T}}
\newcommand{\F}{{\mathcal F}}

\usepackage{bbm} 
\usepackage{mathtools}

\usepackage{cuted}
\usepackage{multirow}

\usepackage{comment}

\usepackage[normalem]{ulem}

\usepackage{subfig}
\icmltitlerunning{Group Equivariant Fourier Neural Operators for Partial Differential Equations}

\begin{document}

\twocolumn[
\icmltitle{Group Equivariant Fourier Neural Operators for Partial Differential Equations}



\icmlsetsymbol{equal}{*}

\begin{icmlauthorlist}
\icmlauthor{Jacob Helwig}{CS,equal}
\icmlauthor{Xuan Zhang}{CS,equal}
\icmlauthor{Cong Fu}{CS}
\icmlauthor{Jerry Kurtin}{CS}
\icmlauthor{Stephan Wojtowytsch}{SW}
\icmlauthor{Shuiwang Ji}{CS}
\end{icmlauthorlist}

\icmlaffiliation{CS}{Department of Computer Science \& Engineering, Texas A\&M
University, TX, USA}
\icmlaffiliation{SW}{Department of Mathematics, Texas A\&M
University, TX, USA}

\icmlcorrespondingauthor{Shuiwang Ji}{sji@tamu.edu}

\icmlkeywords{Machine Learning, ICML}

\vskip 0.3in
]



\printAffiliationsAndNotice{\icmlEqualContribution} 

\begin{abstract}
We consider solving partial differential equations (PDEs) with Fourier neural operators (FNOs), which operate in the frequency domain. Since the laws of physics do not depend on the coordinate system used to describe them, it is desirable to encode such symmetries in the neural operator architecture for better performance and easier learning. While encoding symmetries in the physical domain using group theory has been studied extensively, how to capture symmetries in the frequency domain is under-explored. In this work, we extend group convolutions to the frequency domain and design Fourier layers that are equivariant to rotations, translations, and reflections by leveraging the equivariance property of the Fourier transform. The resulting $G$-FNO architecture generalizes well across input resolutions and performs well in settings with varying levels of symmetry. Our code is publicly available as part of the AIRS library (\url{https://github.com/divelab/AIRS}).
\end{abstract}

    \section{Introduction}

Partial differential equations (PDEs) are widely used to model physical processes that evolve in time and space, including fluid flows~\citep{wang2020towards,bonnet2022airfrans,eckert2019scalarflow}, heat transfer~\citep{zobeiry2021physics} and electromagnetic waves~\citep{lim2022maxwellnet}. Classically, solving a PDE has been viewed as the task of finding a sufficiently smooth function that satisfies a pointwise relationship between derivatives of a different order. A more modern approach is to consider differential operators as (often non-linear) maps between function spaces and utilize techniques of functional analysis to construct and analyze solutions.  The first philosophy is present in neural PDE solvers such as physics-informed neural networks (PINNs) \cite{raissi2019physics, lu2021deepxde}, whereas the second is pursued by neural operators~\citep{lu2021learning, li2021fourier}. While PINNs are used to solve equations individually and online, neural operators learn a solution map between function spaces from problem data to the solution offline. The second approach is highly efficient in contexts where the same problem has to be solved often with slightly varied parameters or initial conditions and is the focus of this work.

\begin{figure}[t]
\begin{center}
\centerline{\includegraphics[width=\columnwidth]{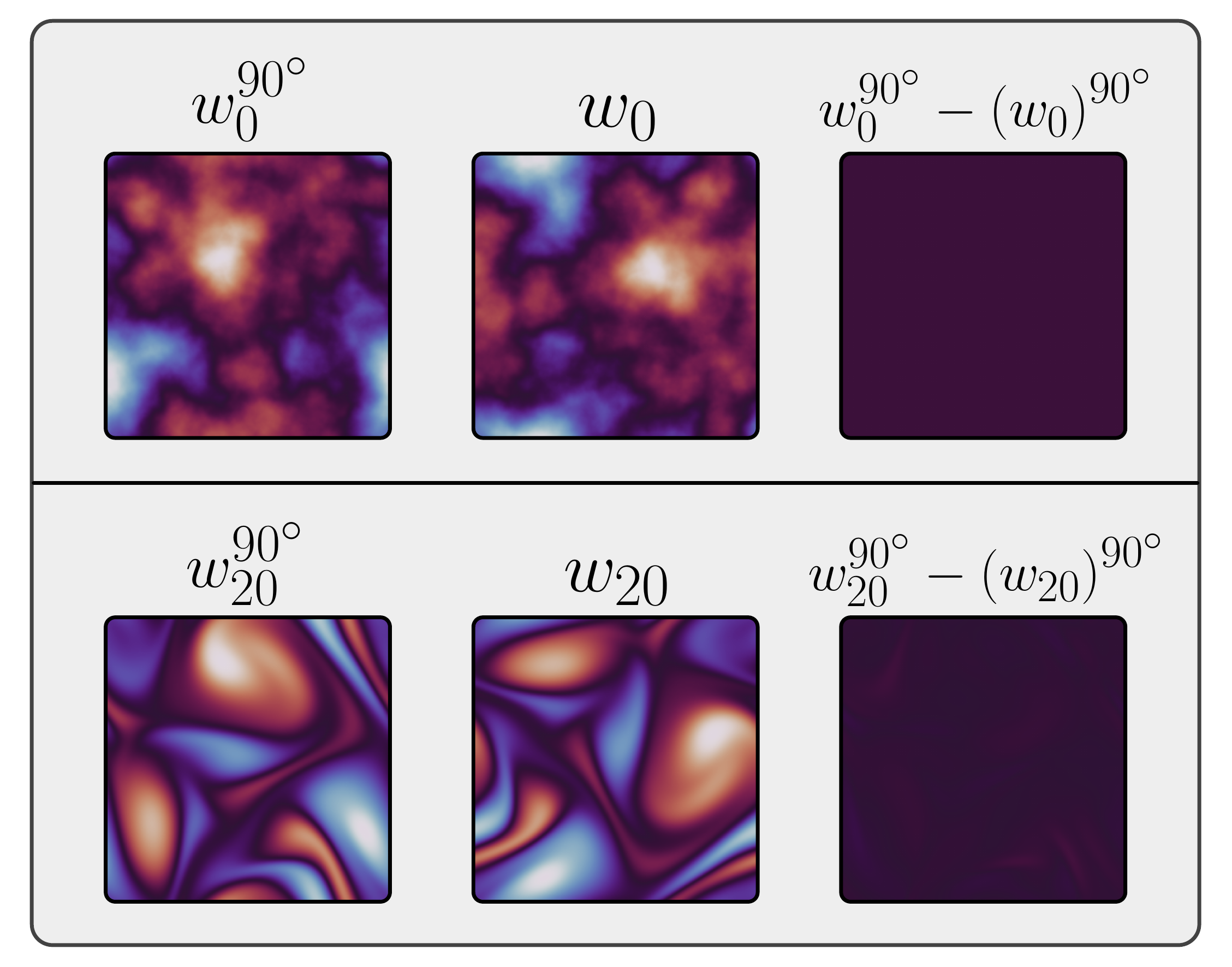}}
\caption{Demonstration of symmetries in the Navier-Stokes equations. We numerically solve the Navier-Stokes equations with a symmetric forcing term as studied in \cref{sec:results}. We find that rotating the initial vorticity field corresponds to a rotated solution vorticity at time $t=20$ up to numerically introduced artifacts, as seen in the rightmost column. We formally derive this symmetry in \cref{app:ns_close}.}
\label{rot_equiv}
\end{center}
\vskip -0.2in
\end{figure}

PDEs capture dynamics of physical processes in which symmetries exist, as visualized in~\cref{rot_equiv}. Symmetries of the underlying problem are reflected in the PDE and its solution operator: the laws of physics do not depend on the coordinate system used to describe them. Many differential operators are rotation invariant, including common models for fluid flow, heat propagation and electrodynamics, but asymmetries in the domain of computation can break symmetries in a more global way: for instance, two directions in a cylindrical container behave similarly while the third plays a different role. It is therefore {\em global} symmetries that solution operators capture. Explicit encoding of these symmetries in network architectures can improve model generalization, interpretability, and sample complexity \citep{weiler2019general, worrall2019deep}. 
           
While equivariant architectures have been studied in diverse applications~\citep{thomas2018tensor, cohen2017steerable, weiler20183d, cohen2018spherical}, most current studies parameterize their convolution kernels in physical space or group space, as opposed to Fourier space. Therefore, the networks are constrained to the resolution of the training data. That is, the trained models may not generalize well to data sampled on a discretization differing from that used in the training data. Additionally, the kernel is most often assumed to have compact, local support, which is effective for sharing information at short distances but requires deep architectures for long-range signal propagation, whereas Fourier convolutions offer an efficient approach for performing global convolutions~\cite{li2021fourier}. 

In this work, we propose the Group Equivariant Fourier Neural Operator ($G$-FNO). By leveraging symmetries of the Fourier transform, we extend group convolution to the frequency domain and design Fourier layers that are equivariant to rotations, reflections and translations. As a result, the proposed $G$-FNO leverages symmetries, can generalize across discretizations, and performs a global convolution that efficiently processes information on multiple scales. Experiments show that $G$-FNO significantly improves the accuracy of PDE solutions even under imperfect symmetries and can generalize to a higher resolution at test time.

\section{Related Work}

\subsection{Neural PDE Solvers}

Neural PDE solvers use neural networks to solve PDEs. Physics-informed neural networks (PINNs)~\citep{raissi2019physics} parameterize solution functions with neural networks and optimize network parameters to satisfy the constraints imposed by a PDE. In contrast, neural operators~\citep{lu2021learning, li2021fourier} learn mappings between input and output functions directly from training data and can be trained autoregressively~\cite{li2021fourier, brandstetter2022message}. As their combination, physics-informed neural operators~\citep{wang2021learning, li2021physics} make use of explicit PDE constraints to help neural operators better satisfy underlying physics.  \citet{bar2019learning, um2020solver, kochkov2021machine} integrate neural solvers and classical solvers~\citep{holl2020phiflow} by using trained neural networks to reduce the numerical error on coarse grids. Our work builds upon Fourier Neural Operators~\citep{li2021fourier, JMLR:v24:21-1524}, which have shown promising results in solving PDEs.

\subsection{Fourier Neural Operator}
\label{sec:FNO_related_work}

Fourier neural operators (FNOs)~\citep{li2021fourier, li2020neural, li2022fourier, JMLR:v24:21-1524} learn to solve PDEs by performing global convolutions via Fourier layers, which are implemented efficiently in the frequency domain using the Fast Fourier Transform (FFT). FNOs process local and global information in parallel through high and low frequency modes~\citep{gupta2022towards}, and reduce computational cost by truncating the highest frequency modes to zero. Additionally, since the convolution operator is learned in the frequency domain, the network is theoretically independent of the resolution of the training data, enabling FNOs to generalize to higher resolution during testing, a task termed \textit{zero-shot super-resolution}~\citep{li2021fourier, boussifmagnet}. 

FNOs have appeared in a variety of applications for dynamics modeling, including optimal control~\citep{hwang2022solving}, modeling of coastal dynamics~\citep{jiang2021digital}, modeling of turbulent Kolmogorov flows~\citep{lilearning}, solving stochastic differential equations~\citep{salvi2022neural}, and forecasting global weather trends~\citep{pathak2022fourcastnet}. Beyond dynamics modeling, \citet{guibas2021adaptive} use Fourier layers to replace spatial self-attention for computer vision tasks.

The FNO architecture has also been extended in recent studies. \citet{poli2022transform} propose a new weight initialization scheme and improved efficiency by only applying one Fourier transform per forward propagation. \citet{tran2021factorized} enable deeper stacks of Fourier layers by applying transforms independently along each of the spatial axes of the input and by proposing a new training strategy. Instead of network design and training, our work focuses on integrating symmetries into FNO architectures by extending group equivariant convolutions to the frequency domain. Specifically, the proposed method parameterizes convolution kernels in the Fourier-transformed group space, and in doing so, allows for a global convolution operator that is equivariant to rotations, reflections, and translations, and can furthermore perform zero-shot super-resolution.

\section{Methods}   

Fourier Neural Operators (FNOs) learn operators mapping an input function to the solution function. For example, for time-dependent PDEs, the input function could be the solution at the current time step, and the output could be the solution at the next time step.

Inspired by the Green's function representation of PDE solutions, FNO alternates between the application of a fixed non-linear map and learned integral operators $\mathcal{K}$, defined as ($\mathcal{K} v)(x) = \int \kappa(x,y) v(y) \text{d}y$, where $\kappa$ is the Green's function to be learned. By further assuming $\kappa$ to be invariant to translations, we can rewrite $\kappa$ as ${\kappa(x,y)=\psi(x - y)}$. Consequently, the operator defines a convolution in physical space as
\begin{equation}\label{Greens_conv}
    (\mathcal{K}v)[x]=\int \psi(x-y)v(y)\text{d}y.
\end{equation}
By the Convolution Theorem, convolution in physical space can be efficiently implemented as element-wise multiplication in the frequency domain, which gives
\begin{equation}\label{op_learn}
    (\mathcal{K}v)[x]=\mathcal F^{-1}\left(\mathcal F\psi\cdot\mathcal Fv\right)[x],
\end{equation}
where $\mathcal{F}$ and $\mathcal{F}^{-1}$ are the Fourier transform and the inverse Fourier transform. Instead of learning the kernel $\psi$, FNO directly learns the Fourier transform of $\psi$, $\mathcal{F}\psi$.

For a translation-invariant Green's function to exist, a PDE must have two properties: linearity and translation-invariance. In practice, most PDEs of interest are non-linear, which FNOs address by including non-linear operations in the learned solution operator. For models in physical sciences, translation-invariance corresponds to the homogeneity of physical space. The assumption is violated in models of heterogeneous materials or boundary phenomena, but generally applies to many situations in fluid mechanics and beyond. If the model is additionally isotropic, the Green's function only depends on the distance $\|x-y\|$, not the direction of $x-y$. In such situations, it is sensible to use computational models which are based on the Green's function representation and respect the invariances it encodes, but allow for non-linear operations when solving non-linear PDEs.

\subsection{Encoding Symmetries in the Physical Domain}

Equivariant architectures have previously been realized using group convolutions ($G$-convolutions) via physically-parameterized kernel functions~\citep{cohen2016group}. It is well-known that convolutional neural networks (CNNs) achieve translation equivariance through convolutional weight-sharing across spatial locations~\citep{lecun1998gradient}. Group equivariant CNNs achieve equivariance to symmetry groups beyond translation by convolving feature maps and kernels defined on these groups. Given a group $G$, $G$-convolutions are defined as
\begin{equation}\label{eq:gconv}
    \left(f\star\psi\right)\left[g\right] = \sum_{h\in G}\sum_{j=1}^{d_z}f^j(h)\left(L_g\psi^j\right)(h),
\end{equation}
where both the feature map $f$ and kernel $\psi$ are functions on $G$ mapping to $\R^{d_z}$, and $L_g\lambda\coloneqq\lambda \circ g^{-1}$ is the group action of $G$ on a function $\lambda$. \citet{cohen2016group} proved the $G$-equivariance of $G$-convolutions. Taking $G=\Z^2$, the group of planar translations, $G$-convolution reduces to the conventional translation-equivariant convolution, i.e., ${\left(f \star \psi\right)(x) = \sum_{y\in \mathbb Z^2}\sum_{j=1}^{d_z} f^j(y)\psi^j(y - x)}$, where ${\left(L_x\psi\right)(y)\coloneqq\psi(y - x)}$ for $x\in\Z^2$.

For the group of translations and $90^\circ$ rotations (the group $p4$), a single-channel $G$-convolution kernel is efficiently implemented as a stack of four independent filters defined on $\mathbb Z^2$, each representing a rotation by $90^\circ$. Applying the group action $L_g$ to the kernel, as in~\cref{eq:gconv}, applies a roto-translation to each of the filters and cyclically shifts their order in the stack. The group $p4m$ additionally considers reflection symmetries and increases the number of independent filters to 8. We review how these groups transform functions defined on $G$ and $\Z^2$ in more depth in~\cref{app:group_intro}.

In the context of solving PDEs, \citet{wang2021incorporating} apply equivariant models leveraging a range of symmetries for modeling the temporal evolution of the velocity field of ocean currents. This work was later extended to settings where dynamics are only approximately equivariant by relaxing the symmetry constraint on learned kernels~\citep{wang2022approximately}. However, in both cases, the convolution kernel was parameterized in physical space, tying the network to the resolution of the training data and lacking parallel processing of multiscale information as is inherent to Fourier convolutions  \cite{gupta2022towards}. \citet{cohen2018spherical} explored parameterizing convolution kernels in the frequency domain to achieve a rotation equivariant convolution for spherical functions. However, this approach is not applicable beyond functions defined on the sphere, and furthermore does not encode translation equivariance, as this is not a symmetry of the sphere. While~\citet{cohen2018spherical} are able to leverage an $SO(3)$ Fourier transform to perform $SO(3)$ equivariant convolutions for spherical functions, no such transform exists for performing convolutions that are equivariant to roto-reflections and translations, which presents a challenge for performing $G$-convolutions in the frequency domain.

        \subsection{$O(2)$-Equivariance of Fourier Transforms}

        To gain insight into how we may perform $G$-convolutions in the frequency domain, we look to characterize the behavior of the Fourier transform of a kernel function $\psi:\mathbb Z^2\to\mathbb R$ under the action of $G$ on $\psi$. In this work, we consider two groups in particular: the group $p4$ generated by $90^\circ$ rotations and translations, and the group $p4m$ generated by $p4$ and horizontal reflections about the origin. We observe an intuitive symmetry, which we prove in \cref{app:four_sym}, that is foundational to performing group convolutions in the frequency domain.

            \begin{lemma}\label{lem:four_sym}
                Given the orthogonal group $O(d)$ acting on functions defined on $\R^d$ by the map $(g,f)\mapsto L_gf$ where $(L_gf)(x) \coloneqq f(g^{-1}x)$, the group action commutes with the Fourier-transform, i.e.\ $\F\circ L_g = L_g\circ \F$.
            \end{lemma}

        This result describes the equivariance of the Fourier transform. That is, \textit{applying a transformation from $O(d)$ to a function in physical space applies the transformation equally to the Fourier transform of the function}. We next use this result for the group of planar roto-reflections $O(2)$ to construct our $G$-Fourier layer.

    \subsection{Group Equivariant Fourier Layers}\label{sec:GrEqFLayers}

        \begin{figure*}[ht]
        \begin{center}
        \centerline{\includegraphics[width=\textwidth]{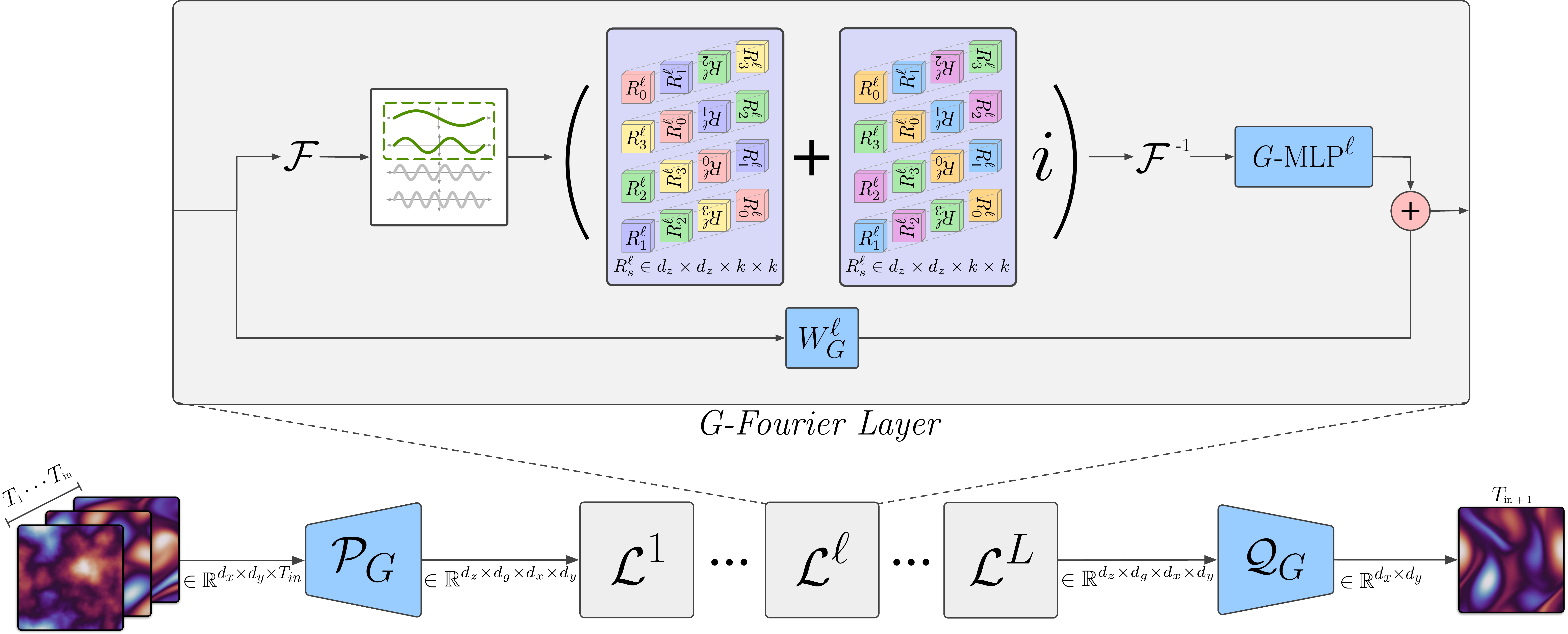}}
        \caption{$G$-FNO-$p4$ architecture, illustrated in the 2D autoregressive form. Bottom: The model raises the input field to a high-dimensional embedding in group space ($\mathcal P_G$) and performs group convolutions in the frequency domain ($\mathcal L^\ell$) before lowering the $G$-feature map back to the base space ($\mathcal Q_G$). Top: A single $G$-Fourier layer. We perform $G$-convolutions in the frequency domain, selecting the $k$ lowest Fourier modes of the input signal. The top row of the real and imaginary parts of the kernel bank illustrate the unrotated kernels, with $R_s^\ell$ rotated by $s\cdot90^\circ$ relative to its canonical orientation, $s\in\{0,1,2,3\}$. \emph{G}-MLP$^\ell$ is a shallow 2-layer MLP with \texttt{GeLU} activation and $1 \times 1$ $G$-Conv layers which we apply along the channel dimension of the output. We add a residual connection linearly projected along the channel dimension by $W_G^\ell$.
        }
        \label{fig:gfno_arch}
        \end{center}
        \vskip -0.2in
        \end{figure*}

        To derive our $G$-Fourier layers, we use the Convolution Theorem in addition to~\cref{lem:four_sym}. For the functions ${v,\rho:\Z^2\to\mathbb R^{d_z}}$ and the translation $x_g\in\mathbb Z^2$, this theorem gives that 
        \begin{align}\label{eq:conv_thm}
            \begin{split}
                \left(v \star \rho\right)\left[x_g\right] & \coloneqq \sum_{x\in\Z^2}\sum_{j=1}^{d_z}v^{j}\left(x\right)\left(L_{x_g}\rho^{j}\right)\left(x\right)
                \\
                & = \sum_{j=1}^{d_z}\mathcal{F}^{-1}\left(\mathcal Fv^{j}\cdot\mathcal F\rho^{j}\right)\left[x_g\right].
            \end{split}
        \end{align}
        The multiplication in the second line is element-wise and $\mathcal F$ is the Discrete Fourier Transform (DFT), which operates on functions defined on $\Z^d$. In contrast, $G$-convolutions operate on functions defined on a group as ${f,\psi:G\to\R^{d_z}}$, for which $\mathcal F$ is not defined in general. However, the groups $G$ we consider here admit a decomposition, as they are the semidirect product $G=\Z^2\rtimes S_G$ of the group of translations $\Z^2$ and the stabilizer of $G$, $S_G$~\cite{weiler2019general}. The stabilizer $S_G$ is defined as transformations that leave the origin invariant, such as rotations for the group $p4$ and rotation-reflections for $p4m$. Therefore, for all group elements $g\in G$, $g$ may be decomposed into a translation $x_g\in\Z^2$ and transformation $s_g\in S_G$ as $g=x_gs_g$, with the action of $G$ similarly decomposed as $L_g=L_{x_g}L_{s_g}$. As shown in the following, this decomposition is key for parameterizing our $G$-convolution kernels in the $\mathcal F$-transformed group space.  
        
        For a fixed stabilizer element $s\in S_G$, define ${f_s:\Z^2\to\R^{d_z}}$ for all translations $x\in\Z^2$ as ${f_s(x)\coloneqq f\left(xs\right)}$, and define ${\psi_s:\mathbb Z^2\to\R^{d_z}}$ analogously using the kernel $\psi$. Then, for the group element ${g=x_gs_g\in G}$, our $G$-Fourier layer is derived from $G$-convolutions as
        \begin{align}\label{eq:g_four_lay}
             & \left(f \star \psi\right)\left[g\right]  \coloneqq \sum_{h\in G}\sum_{j=1}^{d_z}f^{j}\left(h\right)\left(L_g\psi^{j}\right)\left(h\right)\nonumber
             \\
             &\qquad = \sum_{s\in S_G}\sum_{j=1}^{d_z}\sum_{x\in \Z^2}f_{s}^j\left(x\right)\left(L_{x_g}L_{s_g}\psi_{\hat s}^j\right)\left(x\right)\nonumber
             \\
             & \qquad = \sum_{s\in S_G}\sum_{j=1}^{d_z}\mathcal F^{-1}\left(\mathcal Ff_{s}^j\cdot \left(L_{s_g}R_{\hat s}^j\right)\right)\left[x_g\right]
             \\
            & \qquad \eqqcolon \F^{-1}\left(\mathcal Ff\cdot\left(\hat{L}_{s_g}R\right)\right)\left[x_g\right] \nonumber ,
        \end{align}
        where $R_{\hat s}^j\coloneqq\mathcal F\psi_{\hat s}^j$ is the complex-valued function we aim to learn and the equality in~\cref{eq:g_four_lay} is a result of applying the Convolution Theorem in~\cref{eq:conv_thm} followed by~\cref{lem:four_sym}. Additionally, $\hat s\coloneqq s_g^{-1}s$ accounts for the cyclic shift along the stabilizer dimension (absorbed into $\hat{L}_{s_g}$) that we detail in~\cref{app:group_intro}. We furthermore define the order of operations in the action of $S_G$ on $\psi_{\hat{s}}$ as $(L_{s_g}\psi_{\hat{s}})(x)=\psi_{\hat{s}}(s_g^{-1}x)$, as opposed to $(L_{s_g}\psi_{\hat{s}})(x)=(L_{s_g}\psi)(x\hat{s})$, and similarly for $L_{s_g}R_{\hat{s}}$. This ensures that $S_G$, which is a subgroup of $O(2)$, is acting on a function defined on $\Z^2$ via $L_{s_g}$, enabling the use of~\cref{lem:four_sym}. Notably, \cref{eq:g_four_lay} shows that we can efficiently perform group equivariant convolutions in the frequency domain by transforming $R_{s}^j$ by elements $s_g$ from the stabilizer $S_G$.
    
        Concretely, the $\ell$-th $G$-Fourier layer in the $G$-FNO architecture $\mathcal L^\ell$ maps the feature map $f^\ell\in\mathbb R^{d_z\times d_g\times d_x\times d_y}$ to $f^{\ell + 1}\in\mathbb R^{d_z\times d_g\times d_x\times d_y}$. Here, $d_z$ is the dimension of the latent space, $d_g$ is the number of elements in $S_G$ (i.e., $d_g=4$ for $p4$, $8$ for $p4m$), and $d_x\times d_y$ is the resolution of the input function. Our $G$-convolution kernel bank in the frequency domain is then $R^\ell \in\mathbb C^{d_z\times d_z\times d_g\times d_x\times d_y}$, which we visualize in~\cref{fig:gfno_arch}. To manage complexity, we assume \textit{a priori} all modes above a cutoff frequency $k$ are $0$, and thus, only a subset of size $k\times k$ of the $d_x\times d_y$ Fourier modes are learnable, with the remaining modes fixed at $0$.  
        
        Subsetting $R^\ell$ along the output channel dimension, ${R^{\ell, l}\in\mathbb C^{d_z\times d_g\times d_x\times d_y}}$ represents $\mathcal F\psi^{\ell,l}$ for our implicit kernel function $\psi^{\ell,l}:G\to\R^{d_z}$. From~\cref{eq:g_four_lay}, $\psi^{\ell,l}$ is then convolved in the frequency domain with $f^\ell$ to produce the $l$-th channel of $f^{\ell+1}$ as
        \begin{equation}\label{eq:Gfour_lay_ell}
            f^{\ell+1,l}(g)=\mathcal F^{-1}\left(\mathcal Ff^\ell\cdot \left(\hat{L}_{s_g}R^{\ell,l}\right)\right)[x_g].
        \end{equation}
        We denote the operator mapping $f^\ell$ to $f^{\ell+1}$ over all channels and $g\in G$ by $f^{\ell+1}=\mathcal F^{-1}(\mathcal Ff^\ell\cdot \hat{L}_{S_G}R^\ell)$.
        
        To increase the expressive capacity of our $G$-Fourier layer, we compose each frequency domain $G$-convolution with additional equivariant operations. Here, we note that it is key that these operations are applied only along the channel dimension, as aggregating information in physical space would introduce a dependence on the resolution of the training data, whereby reducing the ability of the $G$-FNO to perform super-resolution. The $\ell$-th $G$-Fourier layer $\mathcal L^\ell$ mapping $f^\ell$ to $f^{\ell+1}$ can then be formally expressed as          
        \begin{align}\label{Gfourier_layer}
            \mathcal L^\ell f^\ell = W^{\ell}_G f^\ell + G\text{-MLP}^\ell\left(\mathcal F^{-1}\left(\mathcal Ff^\ell  \cdot \hat{L}_{S_G} R^\ell\right)\right).
        \end{align}
        Here, $W_G^\ell$ linearly projects the residual connection using a $1\times 1$ $G$-Conv layer, and $G\text{-MLP}^\ell$ is a shallow 2-layer MLP with \texttt{GeLU} activation and with the linear layers replaced with $1\times1$ $G$-Conv layers. In~\cref{app:steer}, we consider a steerable parameterization~\cite{cohen2017steerable} of $R^\ell$.         
                        
        \subsection{Group Equivariant Fourier Neural Operator Architecture and Implementation}\label{sec:imple}
            The $G$-FNO composes an encoder $\mathcal P_G$ with multiple $G$-Fourier layers followed by a decoder $\mathcal Q_G$. Both the encoder and decoder are $1\times 1$ $G$-convs that lift the 2D input field $f^0\in\R^{d_{in}\times d_x\times d_y}$ to a higher dimensional embedding in the group space and vice versa. Thus, the overall architecture, visualized in~\cref{fig:gfno_arch}, is expressed as
            \begin{equation}\label{fig:fno_arch}
             \mathcal Q_G\circ \mathcal L^L\circ\dots\circ\mathcal L^2\circ\mathcal L^1 \circ \mathcal P_G.
            \end{equation}
            The encoder additionally concatenates a positional encoding describing the location in space and possibly time to each of the input grid points. As this grid does not transform with the input, the concatenation would not be equivariant if the grid were to contain the Cartesian coordinates of the input grid points. Therefore, we construct a positional encoding that is rotation and reflection invariant by letting each of the grid points map to the distance of the point from the center of the grid. This gives a positional encoding that is invariant to transformations of the input, hence preserving the overall equivariance of the architecture. We also note that although this positional encoding renders architectures only approximately translation equivariant, we show in~\cref{app:pos_enc} that the performance of both the baseline FNO and $G$-FNO are improved with the inclusion of this grid.

            Two additional challenges in efficiently implementing the $G$-Fourier layer concern the organization of the frequency modes in the DFT and enforcing the Hermitian constraint on $R^\ell$. For simplicity in notation, we consider $\psi:G\to\R$ as our implicit kernel function and let our learnable parameter $R:G\to\mathbb C$ be the DFT of $\psi$, with $\psi_s:\Z^2\to\R$ as $\psi$ for a fixed value of $s\in S_G$ and $R_s:\Z^2\to\mathbb C$ defined identically for $R$. First, for the transform of a \textsc{2D} signal, the DFT algorithm places the origin (i.e., the zero frequency) in the upper left corner, followed by the positive modes and then the negative modes along each of the axes. However, transformations from $S_G$ (rotations or roto-reflections) are most naturally applied to $R_s$ with a centered origin. We therefore parameterize $R_s$ as $\mathcal F_\pi\psi_s$, where $\mathcal F_\pi$ represents the DFT followed by a periodic shift to center the zero-frequency component at the origin. In~\cref{eq:g_four_lay}, we must similarly apply $\mathcal F_\pi$ to the input signal $f^{j}_{s}$ so that the corresponding modes are correctly multiplied with $R_s$, and replace $\mathcal F^{-1}$ with $\mathcal F^{-1}_{\pi^{-1}}$, which is the inverse shift followed by the inverse DFT. 
            
            Second, because our implicit kernel function $\psi_s$ is real-valued, the Fourier transform $R_s$ will be Hermitian. That is, $R_s(\xi_1,\xi_2)=R_s^*(-\xi_1,-\xi_2)$, where $\cdot^*$ denotes complex conjugation and $(\xi_1,\xi_2)\in\Z^2$. It is important to enforce the Hermitian property in learning $R$, as it ensures that the result of the multiplication in the frequency domain will also be Hermitian, which in turn ensures that the inverse transform will be real-valued. To enforce this constraint, we only learn the positive modes along the last axis of $R_s$, since the negative modes, which are necessary in~\cref{eq:g_four_lay} for applying transformations from $S_G$, can be directly inferred from the positive modes using the Hermitian property. Maintaining this property has the added benefit that we may use the real FFT algorithm for the forward and inverse transform, which reduces the cost by a factor of roughly 2 compared to the FFT. Further, ensuring the output of our $G$-Fourier layers is real-valued by enforcing the Hermitian constraint allows us to avoid overhead incurred by complex-valued parameters outside of the frequency domain, i.e., in $W^\ell_G$ and $G\text{-MLP}^\ell$.

            We note that the use of Fourier transforms restricts us to studies on the entire space or periodic domains. Numerical discretizations, as we study here, always reduce to the case of a periodic domain, as the DFT implicitly extends the input periodically. Problems on the whole space can be studied, under suitable assumptions, if the length scale of the domain of interest is sufficiently small compared to the assumed domain of periodicity in a discretization.

        \section{Experiments}



We introduce the datasets we consider in~\cref{sec:dataset}, describe our experimental settings in~\cref{sec:experiemental_setting}, and present results in~\cref{sec:results}. 

\subsection{Datasets}\label{sec:dataset}

\begin{figure}[t]
\begin{center}
\centerline{\includegraphics[width=\columnwidth]{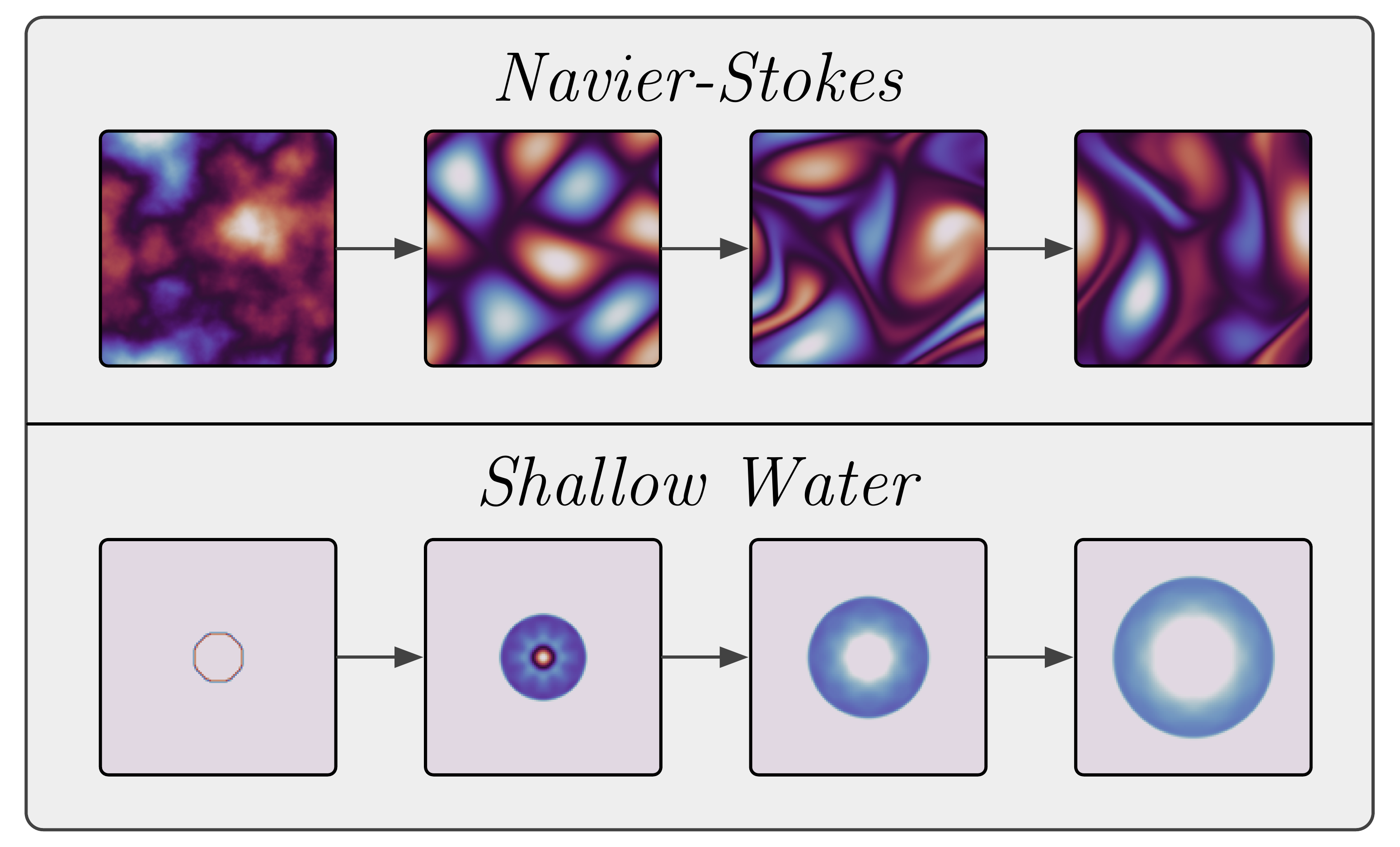}}
\caption{Illustration of the evolution of the Navier-Stokes equations with symmetric forcing (\textsc{NS-sym}, top) and the shallow water equations (\textsc{SWE-sym}, bottom).}
\label{fig:evo}
\end{center}
\vskip -0.2in
\end{figure}

We evaluate our models on two commonly used PDEs in the field of computational fluid dynamics: the incompressible Navier-Stokes equations and the shallow water equations. The Navier-Stokes equations find broad applications in modeling of turbulent dynamics and hydromechanical systems. The shallow water equations are useful in general flooding events simulation~\citep{takamotopdebench} and weather modeling~\cite{gupta2022towards}. 

\textbf{2D Incompressible Navier-Stokes equations. } We consider two versions of the incompressible Navier-Stokes in vorticity form~\cite{li2021fourier}, given by:
\begin{align}
    \label{eq:big_one}
        \partial_t w(x,t) + u(x,t)\cdot\nabla w(x,t)& =\nu\Delta w(x,t) + f(x),
        \\
        \nabla\cdot u(x,t) & = 0, \\
        w(x, 0) & = w_0(x).
\end{align}  
Here, $w(x,t)\in\mathbb R$ is the vorticity field we aim to predict, $w_0(x)\in\mathbb R$ is the initial vorticity, $u(x,t)\in\mathbb R^{2}$ is the velocity, and $\nu=10^{-4}$ is the viscosity coefficient. The solution domain we consider is $x\in[0,1]^2, t\in\{1,2,\ldots, T\}$, where $t$ enumerates a discretization of the continuous temporal domain and $T$ is defined in~\textbf{Tasks and Evaluation}.

In~\cref{eq:big_one}, $f$ is a forcing term that describes external forces acting on the flow. As we show in~\cref{app:ns_close}, for the rotated version of a solution to these equations to still be a solution, the forcing term must be invariant to $90^\circ$ rotations. We therefore consider two realizations of these equations with different forcing terms to evaluate the performance of \textsc{$G$-FNO} in settings with and without global symmetry. The first, studied by~\citet{li2021fourier} and which we refer to as \textsc{NS}, has the forcing term ${f(x_1, x_2)=0.1(\sin(2\pi(x_1+x_2)) + \cos(2\pi (x_1+x_2)))}$, which is not rotation invariant. The second, \textsc{NS-sym}, has the forcing term ${f(x_1, x_2) = 0.1 (\cos(4\pi x_1) + \cos(4  \pi x_2))}$, which is rotation invariant, and hence yields a solution set closed to rotations. We visualize the evolution of the vorticity field $w$ for \textsc{NS-sym} in~\cref{fig:evo,fig:rollout_ns_sym} and for \textsc{NS} in~\cref{fig:rollout_ns_nonsym}. 

\textbf{2D Shallow Water equations. } As with the Navier-Stokes equations, we consider two versions of the shallow water equations that we refer to as \textsc{SWE-sym} and \textsc{SWE}, each with different degrees of global symmetry. \textsc{SWE-sym} is from~\citet{takamotopdebench} and models the dynamics of a fluid that is initially confined within a circular dam and subsequently released due to the sudden removal of the dam. The equations are given by:
\begin{align}
    \partial_t h + \nabla\cdot (hu) & = 0,
    \\
    \partial_t (hu) + \nabla\cdot(huu^\top) + \frac12g\nabla h^2 & = 0,
\end{align}
where $h(x,t)\in\mathbb R$ is the depth that we will predict, $u(x, t)\in\mathbb R^2$ is the velocity, and $g\in\mathbb R$ is acceleration due to gravity. The solution domain is ${x\in[-2.5,2.5]^2}, {t\in\{1,2,\ldots, 25\}}$. The initial depth $h(x, 0)$ is given by:
\begin{equation}\label{eq:init_swe}
    h(x,0)=
    \begin{cases}
        2.0 & r < \sqrt{x_1^2 + x_2^2}
        \\
        1.0 & r \geq \sqrt{x_1^2 + x_2^2}
    \end{cases},
\end{equation}
where $r$ denotes the radial distance to the center of the circular dam. We visualize the evolution of $h$ in~\cref{fig:evo,fig:rollout_SWE_bench}, which demonstrates the high degree of global symmetry present in this data. 

\textsc{SWE} is from~\citet{gupta2022towards} and models both the vorticity field and pressure field of global winds with a large time step size of 48 hours. Because \textsc{SWE} is defined on a rectangular domain, it lacks the global symmetry present in \textsc{SWE-sym}, which is evident from~\cref{fig:rollout_SWE_arena}.

\subsection{Experimental Settings}\label{sec:experiemental_setting}

\textbf{Tasks and Evaluation.}\label{sec:taskeval} We train all models on numerical data that is downsampled from the resolution at which it was generated. In~\cref{sec:generation}, we discuss data generation and downsampling in more depth and include training details in~\cref{sec:training}. We evaluate models on a hold-out test set with the same resolution as the training data. Specifically, for \textsc{NS}, we map the ground truth vorticity field up to $t=10$ to the field at each time step up to $T=30$. Procedures for \textsc{NS-sym} are identical, but with $T=20$ instead. For \textsc{SWE-sym}, we map the depth of the water at $t=1$ up to the depth at $T=25$, while for \textsc{SWE}, we map the pressure and vorticity fields up to $t=2$ to the fields up to $T=11$. In~\cref{sec:roll_vis}, we visualize the predicted rollouts for each dataset. 

We consider two approaches for advancing time with our neural solvers. \textsc{2D} models utilize spatial convolutions with autoregressive predictions, while \textsc{3D} models predict all time steps with a single forward propagation by performing convolutions in space-time. Both \textsc{2D} and \textsc{3D} versions of the equivariant models we consider in our experiments encode symmetries in the spatial domain.

After evaluating rollout errors, we study the equivariance by rotating the test set and evaluating the models again. Additionally, we evaluate models on the super-resolution task in which we apply the models trained on the downsampled training data to the test data at the fine resolution at which it was generated. Note that while the 3D models are able to perform super-resolution in space and time, the 2D models can only perform super-resolution in space, as training the models to advance time autoregressively locks the model to this time step. Lastly, to evaluate the efficacy of super-resolution, we used the predictions for each model made on the coarse grid to interpolate the solutions onto the finer grid.

For each of these tasks, we use relative mean square error to evaluate our models~\cite{li2021fourier,tran2021factorized}. Specifically, 
\begin{align}
    \mathrm{R\textsc{-}MSE} = \frac{1}{n_\text{test}} \sum_{i=1}^{n_\text{test}} \frac{\|\hat{y}_i - y_i\|_2}{\|y_i\|_2},
\end{align}
where $n_\text{test}$ is the number of test PDEs and $\|\cdot\|_2$ is the $L_2$ norm. $\hat{y}_i$ and $y_i$ denote the predicted solution and ground truth of the $i$-th test PDE. We also train our models using the one-step version of this loss, where $\hat y_i$ and $y_i$ are both only one time step. This training strategy was shown by~\citet{tran2021factorized} to be superior to recurrent training, which we further examine in~\cref{app:training_strat}.

\textbf{Baselines. } We consider several variants of the FNO, as well as an equivariant U-Net. 
Coupled with equivariance, this is a particularly interesting comparison, as U-Nets take a sequential approach to processing information on multiple scales, while FNO architectures process multi-scale information in parallel with Fourier convolutions~\cite{gupta2022towards}. 
Our $p4$ equivariant U-Net, \textsc{U-Net-$p4$}, is a modified version of the architecture studied by \citet{wang2021incorporating} for dynamics modeling. The version of the \textsc{FNO} we consider is constructed with a non-equivariant version of our $G$-Fourier layer given in~\cref{Gfourier_layer} and includes versions of the linear projection $W_G^\ell$ and $G\text{-MLP}$ defined for functions on $\mathbb R$ instead of the group space $G$.

We also include versions of the FNO trained using data augmentation to aid the model in learning symmetries in the data, an idea explored for neural solvers by \citet{brandstetter2022lie}. The data augmentation is performed by sampling group transformations from $p4$ (translations and rotations by $90^\circ$) and $p4m$ ($p4$ plus reflections) and applying these transformations to the training data, yielding \textsc{FNO+$p4$} and \textsc{FNO+$p4m$}, respectively. Note that since FNO performs convolution, it is translation equivariant by design, and therefore, we only need to sample rotations and roto-reflections. 

Lastly, we consider \textsc{radialFNO}, a frequency domain parameterization of the architecture explored by~\citet{shen2021rotation}. \citet{shen2021rotation} construct equivariant CNNs using a radial kernel function. By~\cref{lem:four_sym}, the Fourier transform of this kernel will also be radial, and thus, this architecture admits a straightforward extension to a parameterization in the frequency domain. For equivariance to rotations by $90^\circ$, the kernel need not be fully radial and rather just invariant to $90^\circ$ rotations. This increases the capacity of the considered baseline by reducing the degree of weight-sharing required to achieve the desired invariance.

\subsection{Results and Analysis}
\label{sec:results}

\begin{table*}[t]
    \caption{Results on Navier-Stokes with (\textsc{NS-sym}) and without (\textsc{NS}) global symmetry. \textsc{2D} models make rollout predictions autoregressively, while \textsc{3D} models perform convolutions in space-time. We present the relative mean squared error as a percentage averaged over three random seeds for predicted rollouts. Rollouts are of length $10$ in the case of \textsc{NS-sym} and $20$ for \textsc{NS}, and conditioned on the first $10$ time steps in the trajectory for both datasets.}
    \label{tab:NS}
    \vspace{-0.08in}
    \begin{center}
    \begin{sc}
    \resizebox{\textwidth}{!}{
\begin{tabular}{lcccc|cccc}
\toprule
{} & \multicolumn{4}{c}{2D Models} & \multicolumn{4}{c}{3D Models} \\
{} & \multicolumn{2}{c}{NS} & \multicolumn{2}{c}{NS-sym} & \multicolumn{2}{c}{NS} & \multicolumn{2}{c}{NS-sym} \\
{} & \# Par. (M) &                 Test (\%) & \# Par. (M) &                 Test (\%) & \# Par. (M) &                  Test (\%) & \# Par. (M) &                  Test (\%) \\
\midrule
FNO             &      $0.93$ &              $8.41(0.41)$ &      $0.93$ &              $4.21(0.12)$ &      $4.92$ &              $15.84(0.37)$ &      $4.92$ &              $26.02(0.41)$ \\
FNO+$p4$        &      $0.93$ &             $10.44(0.47)$ &      $0.93$ &              $4.80(0.12)$ &      $4.92$ &              $14.14(0.14)$ &      $4.92$ &  $\underline{15.75}(0.33)$ \\
FNO+$p4m$       &      $0.93$ &             $22.09(1.46)$ &      $0.93$ &             $13.06(3.29)$ &      $4.92$ &              $15.32(0.05)$ &      $4.92$ &              $22.25(0.21)$ \\
$G$-FNO-$p4$    &      $0.85$ &     $\mathbf{4.78}(0.39)$ &      $0.85$ &     $\mathbf{2.24}(0.09)$ &      $4.80$ &     $\mathbf{11.77}(0.13)$ &      $4.80$ &     $\mathbf{10.72}(0.27)$ \\
$G$-FNO-$p4m$   &      $0.84$ &  $\underline{6.19}(0.61)$ &      $0.84$ &  $\underline{2.37}(0.19)$ &      $3.89$ &              $12.71(0.31)$ &      $3.89$ &              $17.21(1.35)$ \\
U-Net-$p4$      &      $3.65$ &             $18.40(0.44)$ &      $3.65$ &             $15.39(0.16)$ &      $6.08$ &              $24.62(0.29)$ &      $6.08$ &              $21.82(0.10)$ \\
radialFNO-$p4$  &      $1.03$ &              $9.21(0.26)$ &      $1.03$ &             $12.81(0.42)$ &      $4.98$ &              $12.09(0.08)$ &      $4.98$ &              $17.54(0.60)$ \\
radialFNO-$p4m$ &      $0.95$ &             $10.86(0.18)$ &      $0.95$ &             $17.39(0.22)$ &      $5.63$ &  $\underline{11.83}(0.23)$ &      $5.63$ &              $17.27(0.17)$ \\
\bottomrule
\end{tabular}
    }
    \end{sc}
    \end{center}
    \vspace{-0.05in}
\end{table*}

\begin{table*}[t]
    \caption{Super-resolution results on Navier-Stokes with (\textsc{NS-sym}) and without (\textsc{NS}) global symmetry. We increase the resolution of the test set by a factor of 4 and evaluate models trained on the lower resolution data. For \textsc{2D} models, the super-resolution is in space, while for \textsc{3D} models, the time resolution is also increased. In the column \textsc{SR}, we show the super-resolution rollout errors for models predicting directly to the higher resolution, while \textsc{Int.} shows the error for the predictions made at a lower resolution and fine-grained using interpolation.}
    \label{tab:SRint}
   \vspace{-0.08in}
    \begin{center}
    \begin{sc}
    \resizebox{\textwidth}{!}{
\begin{tabular}{lcccc|cccc}
\toprule
{} & \multicolumn{4}{c}{2D Models} & \multicolumn{4}{c}{3D Models} \\
{} & \multicolumn{2}{c}{NS} & \multicolumn{2}{c}{NS-sym} & \multicolumn{2}{c}{NS} & \multicolumn{2}{c}{NS-sym} \\
{} &               SR Test (\%) &             Int. Test (\%) &               SR Test (\%) &             Int. Test (\%) &               SR Test (\%) &             Int. Test (\%) &               SR Test (\%) &             Int. Test (\%) \\
\midrule
FNO             &     $\mathbf{43.02}(0.18)$ &     $\mathbf{43.14}(0.19)$ &              $32.45(1.47)$ &  $\underline{23.33}(0.07)$ &              $29.99(0.26)$ &              $27.97(0.10)$ &              $31.24(0.66)$ &              $29.38(0.47)$ \\
FNO+$p4$        &              $49.78(8.40)$ &              $43.72(0.45)$ &              $31.72(1.55)$ &     $\mathbf{23.32}(0.09)$ &              $30.36(0.18)$ &  $\underline{27.27}(0.11)$ &              $25.25(0.80)$ &  $\underline{21.20}(0.31)$ \\
FNO+$p4m$       &              $54.04(4.52)$ &              $46.30(1.33)$ &              $32.68(0.84)$ &              $25.02(1.24)$ &              $30.45(0.37)$ &              $27.82(0.12)$ &              $31.44(1.20)$ &              $26.64(0.32)$ \\
$G$-FNO-$p4$    &  $\underline{43.41}(0.12)$ &  $\underline{43.51}(0.11)$ &     $\mathbf{21.89}(0.05)$ &              $23.36(0.04)$ &     $\mathbf{29.62}(0.15)$ &     $\mathbf{27.09}(0.06)$ &     $\mathbf{20.44}(0.73)$ &     $\mathbf{17.71}(0.09)$ \\
$G$-FNO-$p4m$   &              $43.78(0.33)$ &              $43.88(0.30)$ &  $\underline{22.09}(0.03)$ &              $23.56(0.03)$ &              $30.02(0.31)$ &              $27.38(0.21)$ &  $\underline{23.98}(1.30)$ &              $22.14(0.96)$ \\
U-Net-$p4$      &              $92.00(7.22)$ &              $43.68(0.82)$ &              $70.42(1.66)$ &              $24.24(0.12)$ &            $114.99(33.94)$ &              $30.11(0.41)$ &              $79.08(3.60)$ &              $25.85(0.03)$ \\
radialFNO-$p4$  &              $43.73(0.07)$ &              $43.86(0.07)$ &              $25.47(0.31)$ &              $26.52(0.35)$ &  $\underline{29.92}(0.65)$ &              $27.52(0.48)$ &              $24.87(0.51)$ &              $22.83(0.42)$ \\
radialFNO-$p4m$ &              $43.91(0.69)$ &              $44.02(0.71)$ &              $27.85(0.22)$ &              $28.70(0.05)$ &              $29.94(0.60)$ &              $27.40(0.45)$ &              $24.49(0.14)$ &              $22.60(0.14)$ \\
\bottomrule
\end{tabular}
    }
    \end{sc}
    \end{center}
    \vspace{-0.05in}
\end{table*}            

\begin{table*}[t]
    \caption{Rotation test results on Navier-Stokes with (\textsc{NS-sym}) and without (\textsc{NS}) global symmetry. We use models trained on the unrotated data to predict rollouts for the rotated PDE. In the case of \textsc{NS}, rotating the PDE does not produce a solution to the original equations as it does for \textsc{NS-sym}, which we prove in~\cref{app:ns_close}.}
    \label{tab:testrt}
   \vspace{-0.08in}
    \begin{center}
    \begin{sc}
    \resizebox{\textwidth}{!}{
\begin{tabular}{lcccc|cccc}
\toprule
{} & \multicolumn{4}{c}{2D Models} & \multicolumn{4}{c}{3D Models} \\
{} & \multicolumn{2}{c}{NS} & \multicolumn{2}{c}{NS-sym} & \multicolumn{2}{c}{NS} & \multicolumn{2}{c}{NS-sym} \\
{} &                 Test (\%) & Test\textsubscript{$90^\circ$} (\%) &                 Test (\%) & Test\textsubscript{$90^\circ$} (\%) &                  Test (\%) & Test\textsubscript{$90^\circ$} (\%) &                  Test (\%) & Test\textsubscript{$90^\circ$} (\%) \\
\midrule
FNO             &              $8.41(0.41)$ &                      $129.21(3.90)$ &              $4.21(0.12)$ &                        $9.91(0.90)$ &              $15.84(0.37)$ &                      $100.75(2.20)$ &              $26.02(0.41)$ &                       $26.75(0.64)$ \\
FNO+$p4$        &             $10.44(0.47)$ &                       $10.38(0.38)$ &              $4.80(0.12)$ &                        $4.74(0.20)$ &              $14.14(0.14)$ &                       $14.21(0.15)$ &  $\underline{15.75}(0.33)$ &           $\underline{15.85}(0.31)$ \\
FNO+$p4m$       &             $22.09(1.46)$ &                       $22.61(1.54)$ &             $13.06(3.29)$ &                       $12.81(2.80)$ &              $15.32(0.05)$ &                       $15.37(0.03)$ &              $22.25(0.21)$ &                       $22.24(0.20)$ \\
$G$-FNO-$p4$    &     $\mathbf{4.78}(0.39)$ &               $\mathbf{4.78}(0.39)$ &     $\mathbf{2.24}(0.09)$ &               $\mathbf{2.24}(0.09)$ &     $\mathbf{11.77}(0.13)$ &              $\mathbf{11.77}(0.13)$ &     $\mathbf{10.72}(0.27)$ &              $\mathbf{10.72}(0.27)$ \\
$G$-FNO-$p4m$   &  $\underline{6.19}(0.61)$ &            $\underline{6.19}(0.61)$ &  $\underline{2.37}(0.19)$ &            $\underline{2.37}(0.19)$ &              $12.71(0.31)$ &                       $12.71(0.31)$ &              $17.21(1.35)$ &                       $17.21(1.35)$ \\
U-Net-$p4$      &             $18.40(0.44)$ &                       $18.40(0.44)$ &             $15.39(0.16)$ &                       $15.39(0.16)$ &              $24.62(0.29)$ &                       $24.62(0.29)$ &              $21.82(0.10)$ &                       $21.82(0.10)$ \\
radialFNO-$p4$  &              $9.21(0.26)$ &                        $9.21(0.26)$ &             $12.81(0.42)$ &                       $12.81(0.42)$ &              $12.09(0.08)$ &                       $12.09(0.08)$ &              $17.54(0.60)$ &                       $17.54(0.60)$ \\
radialFNO-$p4m$ &             $10.86(0.18)$ &                       $10.86(0.18)$ &             $17.39(0.22)$ &                       $17.39(0.22)$ &  $\underline{11.83}(0.23)$ &           $\underline{11.83}(0.23)$ &              $17.27(0.17)$ &                       $17.27(0.17)$ \\
\bottomrule
\end{tabular}
    }
    \end{sc}
    \end{center}
    \vspace{-0.05in}
\end{table*}            

\begin{table*}[t]
    \caption{Results on Shallow Water Equations with (\textsc{SWE-sym}) and without (\textsc{SWE}) global symmetry. Rollouts are of length $24$ and conditioned on the first time step in the trajectory in the case of \textsc{SWE-sym}. For \textsc{SWE}, rollouts are of length $9$ and conditioned on the first $2$ time steps in the trajectory.}
    \label{tab:SWE}
    \vspace{-0.08in}
    \begin{center}
    \begin{sc}
    \resizebox{\textwidth}{!}{
\begin{tabular}{lcccc|cccc}
\toprule
{} & \multicolumn{4}{c}{2D Models} & \multicolumn{4}{c}{3D Models} \\
{} & \multicolumn{2}{c}{SWE} & \multicolumn{2}{c}{SWE-sym} & \multicolumn{2}{c}{SWE} & \multicolumn{2}{c}{SWE-sym} \\
{} & \# Par. (M) &                  Test (\%) & \# Par. (M) &    Test ($\times10^{-3}$) & \# Par. (M) &                  Test (\%) & \# Par. (M) &    Test ($\times10^{-3}$) \\
\midrule
FNO             &      $6.56$ &              $18.45(0.89)$ &      $0.93$ &              $1.22(0.07)$ &     $49.57$ &     $\mathbf{41.49}(0.12)$ &      $4.92$ &              $1.59(0.02)$ \\
FNO+$p4$        &      $6.56$ &              $33.08(0.20)$ &      $0.93$ &              $1.33(0.08)$ &     $49.57$ &              $44.16(0.14)$ &      $4.92$ &              $1.64(0.02)$ \\
FNO+$p4m$       &      $6.56$ &              $44.24(1.63)$ &      $0.93$ &              $1.32(0.07)$ &     $49.57$ &              $45.06(0.20)$ &      $4.92$ &              $1.65(0.03)$ \\
$G$-FNO-$p4$    &      $6.36$ &     $\mathbf{14.96}(0.06)$ &      $0.85$ &              $1.21(0.02)$ &     $53.70$ &              $43.68(0.55)$ &      $4.80$ &  $\underline{1.44}(0.02)$ \\
$G$-FNO-$p4m$   &      $6.23$ &  $\underline{16.20}(0.17)$ &      $0.84$ &              $1.11(0.17)$ &     $56.81$ &  $\underline{43.46}(0.79)$ &      $3.89$ &     $\mathbf{1.44}(0.02)$ \\
U-Net-$p4$      &      $6.90$ &              $28.79(0.08)$ &      $3.65$ &            $30.86(11.31)$ &      $6.08$ &              $55.28(1.86)$ &      $6.08$ &              $8.03(0.35)$ \\
radialFNO-$p4$  &      $6.79$ &              $23.37(0.14)$ &      $1.03$ &  $\underline{0.71}(0.03)$ &     $53.40$ &              $44.12(0.20)$ &      $4.98$ &              $1.54(0.02)$ \\
radialFNO-$p4m$ &      $6.84$ &              $26.20(0.55)$ &      $0.95$ &     $\mathbf{0.70}(0.07)$ &     $51.92$ &              $44.76(0.15)$ &      $5.63$ &              $1.49(0.01)$ \\
\bottomrule
\end{tabular}
    }
    \end{sc}
    \end{center}
    \vspace{-0.05in}
\end{table*}

We consider two variants of the \textsc{$G$-FNO}: \textsc{$G$-FNO-$p4$}, which employs $G$- Fourier layers that are equivariant to $90^\circ$ rotations and translations, and \textsc{$G$-FNO-$p4m$}, which is additionally equivariant to horizontal and vertical reflections about the origin. In~\cref{app:timemem}, we analyze inference times and forward memory requirements for all models considered.

\subsubsection{Incompressible Navier-Stokes Equations}\label{sec:NS}

\textbf{Test Rollouts.} In~\cref{tab:NS}, we present results on both Navier-Stokes datasets for \textsc{2D} and \textsc{3D} models. $G$-FNO gives the lowest test rollout error in all 4 settings, including improving the baseline error by over $3.5\%$ on the \textsc{NS} data, which lacks global symmetry. In~\cref{app:ghyb}, we examine the effect of only maintaining an equivariant representation in the initial layers of the network on this data, and find that performance increases with the number of equivariant layers. The benefit of equivariant architectures, even on data without global symmetries, has been previously noted~\cite{cohen2016group}, and could be due to the ability of equivariant models to extract local symmetries~\cite{worrall2017harmonic}. By contrast, models trained with data augmentation may achieve an approximate equivariance, but are not constrained to maintain equivariant internal representations and struggle to capture local symmetries~\cite{worrall2017harmonic}. This could explain why the \textsc{FNO} with data augmentation underperforms the \textsc{$G$-FNO} in all settings. This observation may also account for data augmentation reducing the performance of the \textsc{FNO} in all settings except 1, including the \textsc{2D} model on \textsc{NS-sym}, where the global symmetry may deceptively suggest data augmentation as a reasonable choice. 

We also observe that \textsc{U-Net-$p4$} does not perform well in comparison to architectures performing convolutions in the frequency domain. This could be due to the parallel multiscale processing mechanisms inherent to Fourier convolutions, which contrasts the sequential multi-scale processing mechanism employed by U-Nets~\cite{gupta2022towards}. 

\textbf{Super-Resolution.} In~\cref{tab:SRint}, we examine the super-resolution capabilities of $G$-FNO. In 3 out of the 4 settings, $G$-FNO produces the lowest super-resolution error, giving the second-best error to the baseline \textsc{FNO} only for \textsc{2D} models on \textsc{NS}. We also perform super-resolution by fine-graining low-resolution predictions using interpolation. For \textsc{2D} models, we observe for both \textsc{NS} and \textsc{NS-sym} that the direct \textsc{$G$-FNO} super-resolution predictions have a lower error than those made with interpolation. 

\textbf{Rotation Test. } In~\cref{tab:testrt}, we rotate the test data by $90^\circ$ counter-clockwise and evaluate the models trained on the unrotated data. Unsurprisingly, we observe that the errors of all equivariant models are invariant to this transformation. We additionally note that the difference in the rotated test error and unrotated test error on \textsc{NS-sym} is much lower for the \textsc{FNO} than on \textsc{NS}. This empirically demonstrates the low degree of global symmetry present in \textsc{NS} and further emphasizes the ability of $G$-FNO to perform well even in settings lacking such symmetry. 

\subsubsection{Shallow Water equations}

\textbf{Test Rollouts.} In~\cref{tab:SWE}, we present results on both shallow water datasets for \textsc{2D} and \textsc{3D} models. For the \textsc{2D} models on \textsc{SWE-sym}, \textsc{radialFNO} gives the best performance, likely due to the compatibility between the radial kernel and radial solution function shown in~\cref{fig:evo}. We note that as evidenced by the Navier-Stokes experiments and the remaining shallow water settings, the radial inductive bias appears to be overly restrictive and does not generalize well beyond this setting. Beyond \textsc{radialFNO}, \textsc{$G$-FNO} gives the lowest error out of the considered baselines. Additionally, in the \textsc{3D} \textsc{SWE} setting, \textsc{$G$-FNO} is second to the baseline \textsc{FNO}. We note that all models have an error greater than $40\%$ in this setting, suggesting that the \textsc{3D} modeling scheme does not work well for \textsc{SWE}. This could potentially be due to the coarse step size representing 48 hours reducing the smoothness of the function being convolved in space-time.

For \textsc{2D} models on \textsc{SWE} and \textsc{3D} models on \textsc{SWE-sym}, \textsc{$G$-FNO} has the lowest test error. Although \textsc{SWE} lacks global symmetry, which is immediately evident from the non-square domain, \textsc{$G$-FNO} still improves upon the baseline rollout error by over $2\%$. We present super-resolution and rotation test results for \textsc{SWE-sym} in~\cref{app:swe}.
                    
\section{Conclusion}
In this work, we propose to design FNO architectures that encode symmetries. Specifically, by leveraging symmetries of the Fourier transform, we extend group convolutions to the frequency domain and design $G$-Fourier layers that are equivariant to rotations, translations, and reflections. We conduct extensive experiments to evaluate our proposed $G$-FNO. Results show that explicit encoding of symmetries in FNO architectures leads to consistent performance improvements.

\section*{Acknowledgements}

This work was supported in part by National Science Foundation grant IIS-2006861, and by state allocated funds for the Water Exceptional Item through Texas A\&M AgriLife Research facilitated by the Texas Water Resources Institute.
                
\bibliography{main}
\bibliographystyle{icml2023}

\newpage
\appendix
\onecolumn


            \begin{table*}[t]
                \caption{Inference time and forward memory requirements. We analyze the time and space complexity of all models on the \textsc{NS} data over 10,000 forward propagations. As in our experiments, we use a batch size of $20$ for the 2D models trained to make predictions autoregressively, and a batch size of $10$ for the 3D models that perform convolutions in space-time.}
                \label{tab:timemem}
                \begin{center}
                \begin{sc}
                \resizebox{\columnwidth}{!}{
                    \begin{tabular}{l>{\centering\arraybackslash}m{2.5cm}>{\centering\arraybackslash}m{2.5cm}>{\centering\arraybackslash}m{2.5cm}|>{\centering\arraybackslash}m{2.5cm}>{\centering\arraybackslash}m{2.5cm}>{\centering\arraybackslash}m{2.5cm}}
\toprule
{} & \multicolumn{3}{c}{2D Models} & \multicolumn{3}{c}{3D Models} \\
{} & \# Par. (M) & Inference Time (ms) & Forward Memory (GiB) & \# Par. (M) & Inference Time (ms) & Forward Memory (GiB) \\
\midrule
FNO             &              0.93 &          4.27(1.66) &                 1.83 &              4.92 &          8.71(7.54) &                 3.09 \\
$G$-FNO-$p4$    &              0.85 &           4.50(2.12) &                 1.97 &              4.80 &        10.21(15.62) &                 4.48 \\
$G$-FNO-$p4m$   &              0.84 &          4.81(3.79) &                 2.17 &              3.89 &        10.73(16.46) &                 5.41 \\
radialFNO-$p4$  &              1.03 &          4.34(2.63) &                 1.97 &              4.98 &        10.81(16.58) &                 6.24 \\
radialFNO-$p4m$ &              0.95 &          4.16(3.29) &                 2.03 &              5.63 &        11.53(17.98) &                 7.50 \\
U-Net-$p4$      &              3.65 &          7.92(4.34) &                 2.31 &              6.08 &        20.29(15.77) &                 7.45 \\
\bottomrule
\end{tabular}

                                        }
                                        \end{sc}
                \end{center}
                \vskip -0.1in
            \end{table*}            

\begin{table*}[t]
    \caption{\textsc{$G$-Hybrid} results on \textsc{NS}. We let the first $N$ layers of the network be equivariant $p4$ and $p4m$ $G$-Fourier layers, and the remaining $4-N$ layers be non-equivariant Fourier layers. We report both the test and rotated test errors. }
    \label{tab:ghybrid}
    \begin{center}
    \begin{small}
    \begin{sc}
\begin{tabular}{cccc|ccc}
\toprule
                      & \multicolumn{3}{c}{$p4$} & \multicolumn{3}{c}{$p4m$} \\
\# $G$-Fourier Layers & \# Par. (M) &                Test (\%) & Test\textsubscript{$90^\circ$} (\%) & \# Par. (M) &                Test (\%) & Test\textsubscript{$90^\circ$} (\%) \\
\midrule
                    0 &      $0.93$ &             $8.41(0.41)$ &                      $129.21(3.90)$ &      $0.93$ &             $8.41(0.41)$ &                      $129.21(3.90)$ \\
                    1 &      $1.14$ &             $7.17(0.51)$ &          $\underline{126.53}(3.66)$ &      $1.32$ & $\underline{7.04}(0.41)$ &                      $128.32(1.83)$ \\
                    2 &      $1.12$ &             $6.30(0.82)$ &                      $127.89(4.99)$ &      $1.30$ &             $7.32(0.37)$ &                      $129.15(3.02)$ \\
                    3 &      $1.10$ & $\underline{6.12}(0.44)$ &                      $128.73(7.78)$ &      $1.27$ &             $7.29(0.14)$ &          $\underline{125.74}(9.08)$ \\
                    4 &      $0.85$ &    $\mathbf{4.78}(0.39)$ &               $\mathbf{4.78}(0.39)$ &      $0.84$ &    $\mathbf{6.19}(0.61)$ &               $\mathbf{6.19}(0.61)$ \\
\bottomrule
\end{tabular}
    
    \end{sc}
    \end{small}
    \end{center}
    \vskip -0.1in
\end{table*}            

\begin{table*}[t]
    \caption{Super-resolution results on \textsc{SWE-sym}. We increase the resolution of the test set by a factor of 4 and evaluate models trained on the lower resolution data. For \textsc{2D} models, the super-resolution is in space, while for \textsc{3D} models, the time resolution is also increased. In the column \textsc{SR}, we show the super-resolution rollout errors for models predicting directly to the higher resolution, while \textsc{Int.} shows the error for the predictions made at a lower resolution and fine-grained using interpolation.}
    \label{tab:SRint_swe}
    \begin{center}
    \begin{small}
    \begin{sc}
\begin{tabular}{lcc|cc}
\toprule
{} & \multicolumn{2}{c}{2D Models} & \multicolumn{2}{c}{3D Models} \\
{} &  SR Test ($\times10^{-3}$) & Int. Test ($\times10^{-3}$) &  SR Test ($\times10^{-3}$) & Int. Test ($\times10^{-3}$) \\
\midrule
FNO             &  $\underline{15.56}(2.92)$ &               $16.15(0.01)$ &              $16.90(0.58)$ &               $17.76(0.01)$ \\
FNO+$p4$        &     $\mathbf{14.80}(3.30)$ &               $16.16(0.01)$ &              $17.19(0.86)$ &               $17.76(0.01)$ \\
FNO+$p4m$       &              $16.38(3.94)$ &               $16.16(0.01)$ &              $17.11(0.83)$ &               $17.75(0.01)$ \\
$G$-FNO-$p4$    &              $19.39(4.60)$ &               $16.15(0.00)$ &              $15.87(0.60)$ &      $\mathbf{17.75}(0.00)$ \\
$G$-FNO-$p4m$   &             $31.00(11.27)$ &               $16.15(0.01)$ &              $16.88(1.19)$ &   $\underline{17.75}(0.00)$ \\
U-Net-$p4$      &         $3009.65(2348.93)$ &               $35.19(9.48)$ &            $167.76(36.77)$ &               $19.19(0.13)$ \\
radialFNO-$p4$  &              $30.40(4.20)$ &   $\underline{16.12}(0.00)$ &  $\underline{14.48}(0.76)$ &               $17.76(0.00)$ \\
radialFNO-$p4m$ &              $22.42(0.95)$ &      $\mathbf{16.12}(0.00)$ &     $\mathbf{13.36}(0.36)$ &               $17.76(0.00)$ \\
\bottomrule
\end{tabular}
    \end{sc}
    \end{small}
    \end{center}
    \vskip -0.1in
\end{table*}

\begin{table*}[t]
    \caption{Rotation test results for \textsc{SWE-sym}. We use models trained on the unrotated data to predict rollouts for the rotated PDE.}
    \label{tab:swe_rt}
    \begin{center}
    \begin{small}
    \begin{sc}
\begin{tabular}{lcc|cc}
\toprule
{} & \multicolumn{2}{c}{2D Models} & \multicolumn{2}{c}{3D Models} \\
{} &    Test ($\times10^{-3}$) & Test\textsubscript{$90^\circ$} ($\times10^{-3}$) &    Test ($\times10^{-3}$) & Test\textsubscript{$90^\circ$} ($\times10^{-3}$) \\
\midrule
FNO             &              $1.22(0.07)$ &                                     $1.50(0.05)$ &              $1.59(0.02)$ &                                     $1.80(0.02)$ \\
FNO+$p4$        &              $1.33(0.08)$ &                                     $1.34(0.10)$ &              $1.64(0.02)$ &                                     $1.69(0.03)$ \\
FNO+$p4m$       &              $1.32(0.07)$ &                                     $1.33(0.07)$ &              $1.65(0.03)$ &                                     $1.71(0.03)$ \\
$G$-FNO-$p4$    &              $1.21(0.02)$ &                                     $1.21(0.02)$ &  $\underline{1.44}(0.02)$ &                         $\underline{1.44}(0.02)$ \\
$G$-FNO-$p4m$   &              $1.11(0.17)$ &                                     $1.11(0.17)$ &     $\mathbf{1.44}(0.02)$ &                            $\mathbf{1.44}(0.02)$ \\
U-Net-$p4$      &            $30.86(11.31)$ &                                   $30.86(11.31)$ &              $8.03(0.35)$ &                                     $8.03(0.35)$ \\
radialFNO-$p4$  &  $\underline{0.71}(0.03)$ &                         $\underline{0.71}(0.03)$ &              $1.54(0.02)$ &                                     $1.54(0.02)$ \\
radialFNO-$p4m$ &     $\mathbf{0.70}(0.07)$ &                            $\mathbf{0.70}(0.07)$ &              $1.49(0.01)$ &                                     $1.49(0.01)$ \\
\bottomrule
\end{tabular}
    \end{sc}
    \end{small}
    \end{center}
    \vskip -0.1in
\end{table*}            

\begin{table*}[t]
    \caption{Training strategy analysis for \textsc{NS}. We present the best validation result and super-resolution test error for three choices of training strategy: \textsc{Markov}, where we condition on one time step and predict one time step into the future, \textsc{Recurrent}, where we condition on several time steps and predict the remaining steps in the rollout autoregressively, and \textsc{Teacher Forcing}, where we condition on several time steps and predict 1 time step into the future.}
    \label{tab:trainstrat}
    \begin{center}
    \begin{small}
    \begin{sc}
    \begin{tabular}{llcc}
\toprule
                                 &       Strategy          &                 Valid (\%) &                SR Test (\%) \\
\midrule
\multirow{3}{*}{FNO} & Markov &      $\mathbf{7.19}(0.28)$ &               $61.10(2.58)$ \\
                                 & Recurrent &              $16.23(0.49)$ &      $\mathbf{42.14}(0.25)$ \\
                                 & Teacher Forcing &   $\underline{8.64}(0.19)$ &   $\underline{43.02}(0.18)$ \\
\midrule
\multirow{3}{*}{FNO+$p4$} & Markov &  $\underline{10.90}(0.24)$ &               $60.46(2.92)$ \\
                                 & Recurrent &              $17.60(0.14)$ &      $\mathbf{41.22}(0.41)$ \\
                                 & Teacher Forcing &     $\mathbf{10.70}(0.49)$ &   $\underline{49.78}(8.40)$ \\
\midrule
\multirow{3}{*}{FNO+$p4m$} & Markov &     $\mathbf{18.26}(0.75)$ &               $60.41(2.28)$ \\
                                 & Recurrent &  $\underline{20.81}(0.28)$ &      $\mathbf{40.63}(0.20)$ \\
                                 & Teacher Forcing &              $22.43(1.59)$ &   $\underline{54.04}(4.52)$ \\
\midrule
\multirow{3}{*}{$G$-FNO-$p4$} & Markov &   $\underline{5.06}(0.04)$ &   $\underline{43.35}(0.24)$ \\
                                 & Recurrent &              $13.20(0.06)$ &      $\mathbf{42.36}(0.23)$ \\
                                 & Teacher Forcing &      $\mathbf{4.86}(0.32)$ &               $43.41(0.12)$ \\
\midrule
\multirow{3}{*}{$G$-FNO-$p4m$} & Markov &      $\mathbf{6.59}(0.96)$ &   $\underline{43.18}(0.82)$ \\
                                 & Recurrent &              $13.93(0.05)$ &      $\mathbf{42.16}(0.34)$ \\
                                 & Teacher Forcing &   $\underline{6.73}(0.82)$ &               $43.78(0.33)$ \\
\midrule
\multirow{3}{*}{U-Net-$p4$} & Markov &  $\underline{33.70}(4.00)$ &  $\underline{110.60}(0.92)$ \\
                                 & Recurrent &              $51.61(7.68)$ &              $112.43(3.84)$ \\
                                 & Teacher Forcing &     $\mathbf{18.86}(0.53)$ &      $\mathbf{92.00}(7.22)$ \\
\midrule
\multirow{3}{*}{radialFNO-$p4$} & Markov &      $\mathbf{8.43}(0.25)$ &   $\underline{43.14}(0.29)$ \\
                                 & Recurrent &              $13.51(0.07)$ &      $\mathbf{42.55}(0.05)$ \\
                                 & Teacher Forcing &   $\underline{9.59}(0.05)$ &               $43.73(0.07)$ \\
\midrule
\multirow{3}{*}{radialFNO-$p4m$} & Markov &     $\mathbf{10.38}(0.19)$ &   $\underline{43.74}(0.09)$ \\
                                 & Recurrent &              $13.81(0.21)$ &      $\mathbf{42.30}(0.27)$ \\
                                 & Teacher Forcing &  $\underline{11.51}(0.68)$ &               $43.91(0.69)$ \\
\bottomrule
\end{tabular}
    \end{sc}
    \end{small}
    \end{center}
    \vskip -0.1in
\end{table*}            

\begin{table*}[t]
    \caption{Positional encoding analysis for \textsc{NS}. We present the best validation, test, and rotation test results for three choices of positional encoding: \textsc{None}, where we do not encode position, \textsc{Symmetric}, where the encoding for each point is the distance from the middle of the grid, and \textsc{Cartesian}, where the encoding is the Cartesian coordinates of the grid point.}
    \label{tab:posenc}
    \begin{center}
    \begin{small}
    \begin{sc}
\begin{tabular}{llccc}
\toprule
                                 &  Positional Encoding         &                 Valid (\%) &                  Test (\%) & Test\textsubscript{$90^\circ$} (\%) \\
\midrule
\multirow{3}{*}{FNO} & None &               $9.05(0.19)$ &   $\underline{8.54}(0.36)$ &                      $130.14(2.06)$ \\
                                 & Symmetric &   $\underline{9.01}(0.26)$ &               $8.95(0.18)$ &             $\mathbf{129.08}(3.95)$ \\
                                 & Cartesian &      $\mathbf{8.64}(0.19)$ &      $\mathbf{8.41}(0.41)$ &          $\underline{129.21}(3.90)$ \\
\midrule
\multirow{3}{*}{FNO+$p4$} & None &              $10.82(0.21)$ &  $\underline{10.46}(0.31)$ &           $\underline{10.46}(0.46)$ \\
                                 & Symmetric &     $\mathbf{10.63}(0.07)$ &              $11.04(0.26)$ &                       $10.47(0.36)$ \\
                                 & Cartesian &  $\underline{10.70}(0.49)$ &     $\mathbf{10.44}(0.47)$ &              $\mathbf{10.38}(0.38)$ \\
\midrule
\multirow{3}{*}{FNO+$p4m$} & None &  $\underline{23.54}(0.57)$ &  $\underline{23.07}(1.29)$ &                       $23.65(1.04)$ \\
                                 & Symmetric &              $23.67(1.31)$ &              $23.29(0.44)$ &           $\underline{23.09}(0.29)$ \\
                                 & Cartesian &     $\mathbf{22.43}(1.59)$ &     $\mathbf{22.09}(1.46)$ &              $\mathbf{22.61}(1.54)$ \\
\midrule
\multirow{3}{*}{$G$-FNO-$p4$} & None &      $\mathbf{4.45}(0.27)$ &   $\underline{4.47}(0.21)$ &            $\underline{4.47}(0.21)$ \\
                                 & Symmetric &               $4.86(0.32)$ &               $4.78(0.39)$ &                        $4.78(0.39)$ \\
                                 & Cartesian &   $\underline{4.61}(0.25)$ &      $\mathbf{4.39}(0.25)$ &               $\mathbf{4.39}(0.25)$ \\
\midrule
\multirow{3}{*}{$G$-FNO-$p4m$} & None &               $7.15(0.25)$ &               $6.75(0.17)$ &                        $6.75(0.17)$ \\
                                 & Symmetric &      $\mathbf{6.73}(0.82)$ &      $\mathbf{6.19}(0.61)$ &               $\mathbf{6.19}(0.61)$ \\
                                 & Cartesian &   $\underline{6.86}(0.17)$ &   $\underline{6.67}(0.32)$ &            $\underline{6.67}(0.32)$ \\
\midrule
\multirow{3}{*}{U-Net-$p4$} & None &     $\mathbf{18.00}(0.60)$ &  $\underline{17.95}(0.07)$ &              $\mathbf{17.95}(0.06)$ \\
                                 & Symmetric &              $18.86(0.53)$ &              $18.40(0.44)$ &                       $18.40(0.44)$ \\
                                 & Cartesian &  $\underline{18.84}(1.15)$ &     $\mathbf{17.73}(0.13)$ &           $\underline{18.07}(0.27)$ \\
\midrule
\multirow{3}{*}{radialFNO-$p4$} & None &      $\mathbf{9.56}(0.31)$ &      $\mathbf{9.13}(0.09)$ &               $\mathbf{9.13}(0.09)$ \\
                                 & Symmetric &   $\underline{9.59}(0.05)$ &   $\underline{9.21}(0.26)$ &            $\underline{9.21}(0.26)$ \\
                                 & Cartesian &               $9.79(0.31)$ &               $9.58(0.05)$ &                      $24.75(20.20)$ \\
\midrule
\multirow{3}{*}{radialFNO-$p4m$} & None &              $11.89(0.34)$ &  $\underline{11.01}(0.04)$ &           $\underline{11.01}(0.04)$ \\
                                 & Symmetric &  $\underline{11.51}(0.68)$ &     $\mathbf{10.86}(0.18)$ &              $\mathbf{10.86}(0.18)$ \\
                                 & Cartesian &     $\mathbf{11.36}(0.44)$ &              $11.20(0.12)$ &                       $11.84(1.02)$ \\
\bottomrule
\end{tabular}
\end{sc}
    \end{small}
    \end{center}
    \vskip -0.1in
\end{table*}            

\begin{table*}[t]
    \caption{Steerable $G$-FNO results on \textsc{NS}. We present test and rotation test results for a steerable parameterization of the $G$-FNO.}
    \label{tab:steer}
    \begin{center}
    \begin{small}
    \begin{sc}
\begin{tabular}{lccc}
\toprule
{} & \# Par. (M) &                 Test (\%) & Test\textsubscript{$90^\circ$} (\%) \\
\midrule
$G$-FNO-$p4$-steer  &      $0.83$ &             $20.87(1.25)$ &                       $20.87(1.25)$ \\
$G$-FNO-$p4m$-steer &      $0.89$ &             $22.58(0.41)$ &                       $22.58(0.41)$ \\
$G$-FNO-$p4$        &      $0.85$ &     $\mathbf{4.78}(0.39)$ &               $\mathbf{4.78}(0.39)$ \\
$G$-FNO-$p4m$       &      $0.84$ &  $\underline{6.19}(0.61)$ &            $\underline{6.19}(0.61)$ \\
\bottomrule
\end{tabular}
    \end{sc}
    \end{small}
    \end{center}
    \vskip -0.1in
\end{table*}            

    \section{Training and Data Generation Details}

\subsection{Data Generation}\label{sec:generation}

For \textsc{NS-sym}, we generate 1,000 training trajectories, $100$ validation trajectories, and $100$ test trajectories using the psuedo-spectral Crank-Nicolson solver from~\citet{li2021fourier}. For \textsc{NS}, we use the data directly from \citet{li2021fourier}, again with a 1,000/100/100 split. For each trajectory, the boundary conditions are periodic and the initial conditions $w_0(x)$ are sampled from a Gaussian random field. Trajectories were solved numerically on a $256\times 256\times 120$ grid and downsampled to $64\times 64\times 30$, where the first two dimensions are in space and the third is time.

For \textsc{SWE-sym}, we used numerical data from~\citet{takamotopdebench} generated using the finite volume method. We split 1,000 trajectories into $800$ training trajectories, $100$ validation trajectories and $100$ test trajectories. For each trajectory, the radius of the dam, $r$ in~\cref{eq:init_swe}, is sampled uniformly from $(0.3, 0.7)$. We downsample the numerical solution from $128\times 128\times 100$ to $32\times 32\times 25$. Unlike \textsc{NS} and \textsc{NS-sym}, we performed spatial downsampling using $2\times 2$ mean-pooling, as strided downsampling introduced a significant asymmetry that was not present in the original data. 

For \textsc{SWE}, we follow the methods and splits used by \citet{gupta2022towards} to generate 5,600 training trajectories, 1,120 validation trajectories, and 1,120 test trajectories, all with resolution $96\times192\times 88$. We follow~\citet{gupta2022towards} in temporally downsampling by a factor of $8$ to $11$ total time steps spaced $48$ hours apart.

\subsection{Training}\label{sec:training}

We perform 3 replicates of all experiments and closely follow the training strategy and hyperparameter specification scheme of \citet{li2021fourier}. We use 4 Fourier layers for all frequency domain models, truncating the transform to the $12$ lowest Fourier modes for all 2D models and $8$ spatial/$6$ temporal Fourier modes for all 3D models. In the case of \textsc{SWE}, we increase the number of modes for all \textsc{2D} models to $32$ following~\citet{gupta2022towards} and $22$ spatial/$8$ temporal modes for all \textsc{3D} models. We also replace the \textsc{Symmetric} positional encoding discussed in~\cref{app:pos_enc} for \textsc{$G$-FNO} on \textsc{SWE} with the \textsc{Cartesian} encoding, as the non-square $96\times192$ spatial domain prevents the \textsc{Symmetric} encoding from being invariant to rotations, breaking the equivariance of the model.

For the baseline \textsc{FNO}, the dimension of the latent space is $20$. For \textsc{$G$-FNO}, we offset the extra dimensions added to kernels discussed in~\cref{sec:GrEqFLayers} by reducing the number of channels to give a roughly equal number of trainable parameters compared to the baseline \textsc{FNO}. Specifically, in the case of \textsc{2D} models, the dimension of the latent space for \textsc{$G$-FNO-$p4$} and \textsc{$G$-FNO-$p4m$} is $10$ and $7$, respectively. For \textsc{3D} models, the dimensions are $11$ and $7$, except in the case of \textsc{SWE}, where we increase the dimension for \textsc{$G$-FNO-$p4m$} to $8$. We also increased the first-layer latent dimension to 100 for a 48.09M parameter \textsc{3D} \textsc{U-Net-$p4$} on \textsc{SWE}. However, the 6.08M parameter version with latent dimension 32 presented in~\cref{tab:SWE} improves the test error over the larger model by roughly $10\%$.

As opposed to the \textsc{Recurrent} training strategy employed by \citet{li2021fourier} for the 2D models, we instead use the \textsc{Teacher Forcing} strategy, which was shown by \citet{tran2021factorized} to improve performance. In~\cref{app:training_strat}, we compare training strategies on \textsc{NS}. We use the Adam optimizer~\citep{kingma2015adam} with $\beta_1=0.9,\beta_2=0.999$, and weight decay $10^{-4}$. We use batch size of $20$ for \textsc{2D} models and $10$ for \textsc{3D} models with a cosine learning rate scheduler that starts at $10^{-3}$ and is decayed to $0$. \textsc{3D} models are trained for $500$ epochs or until the validation loss does not improve for 100 successive epochs, while \textsc{2D} models are trained for $100$ epochs. \textsc{2D} models are trained for less epochs because each PDE in the training set corresponds to 1 training example for the 3D models and $T-T_{in}$ training examples for the 2D models, where $T_{in}$ is the number of time steps being conditioned on. All models are implemented using PyTorch~\cite{Paszke_PyTorch_An_Imperative_2019} and trained on a single NVIDIA A100 80GB GPU.

    \section{Additional Results}

        \subsection{Inference Time and Forward Memory Requirements}\label{app:timemem}
        
            In~\cref{tab:timemem}, we present the average inference time across 10,000 forward evaluations and the GPU memory utilized for each model on the \textsc{NS} data. Here, we note that the times reported for \textsc{2D} models are for predictions only one step into the future, while \textsc{3D} models perform convolutions in space and time and thus predict all $T=20$ steps with one forward pass. We use batch size of 20 for \textsc{2D} models and \textsc{10} for 3D models.

        \subsection{\textsc{$G$-Hybrid} Experiments on \textsc{NS}}\label{app:ghyb}

            We explore the effect of only maintaining an equivariant representation in the initial layers by composing ${N\in\{0,1,2,3,4\}}$ $G$-Fourier layers and replacing the final $4-N$ layers with non-equivariant FNO layers. We train 2D models to make predictions autoregressively on the \textsc{NS} task described in~\cref{sec:taskeval} and present rollout errors on the test set and rotated test set in~\cref{tab:ghybrid}. 

            We observe that increasing the number of equivariant layers improves performance, despite the \textsc{NS} data not being globally symmetric, as we prove in~\cref{app:ns_close}. Equivariant models have been noted to offer benefits on non-symmetric datasets in the past~\cite{cohen2016group}. The observed benefits could stem from the ability of equivariant layers to learn local symmetries~\cite{worrall2017harmonic}.

    \subsection{\textsc{SWE-sym} Super-Resolution and Rotation Test Results}\label{app:swe}

        In~\cref{tab:SRint_swe,tab:swe_rt}, we present super-resolution and rotation test results for \textsc{SWE-sym}. For the super-resolution task with \textsc{2D} models that make rollout predictions autoregressively, the ground truth rollouts during training are $32\times32\times24$ and $128\times128\times24$ during testing, where the first two dimensions are spatial and the third is the number of time steps. For \textsc{3D} models that perform convolutions in space-time, the temporal resolution also increases during testing to $128\times128\times96$. 
    
    \subsection{Training Strategy for \textsc{2D} Models}\label{app:training_strat}

        We consider three different variants of training for 2D models: \textsc{Recurrent}, \textsc{Teacher Forcing}, and \textsc{Markov}. \citet{li2021fourier} used \textsc{Recurrent} training for their \textsc{2D} FNO, wherein the model predicts the entire rollout during training autoregressively and the loss is back-propagated through time. \citet{tran2021factorized} found that \textsc{Teacher Forcing} and \textsc{Markov} training improved performance of FNO architectures relative to \textsc{Recurrent} training. In both strategies, the model is trained to make predictions only one step into the future conditioned on the ground truth solutions at the previous time steps. However, under the \textsc{Markov} strategy, the model is only conditioned on the most recent time step, whereas models trained with \textsc{Teacher Forcing} are conditioned on several of the most recent time steps. 
        
        In~\cref{tab:trainstrat}, we compare these training strategies based on their validation errors and super-resolution test errors. In all cases, our findings agree with those of~\citet{tran2021factorized} in that \textsc{Markov} and \textsc{Teacher Forcing} improve results relative to \textsc{Recurrent} training. While for some models, \textsc{Markov} training gives a better error than \textsc{Teacher Forcing}, we find that \textsc{Markov} training significantly reduces the ability of the baseline FNO to perform super-resolution, and thus, we opt to train all models using the \textsc{Teacher Forcing} strategy.
        
    \subsection{Positional Encoding}\label{app:pos_enc}

        We consider 3 variants of positional encoding: \textsc{None}, where position is not encoded, \textsc{Symmetric}, where the encoding represents the distance of the grid point from the center of the grid, giving a roto-reflection invariant grid, and \textsc{Cartesian}, where the encoding is the Cartesian coordinates of each of the grid points. The resulting grid is then concatenated to the input of the network. As the grid is fixed and does not transform with the input, \textsc{Symmetric} breaks translation equivariance while preserving roto-reflection equivariance, while \textsc{Cartesian} breaks translation and roto-reflection equivariance. \textsc{None} preserves all symmetries.

        We present validation, test, and rotation test results for these encodings in~\cref{tab:posenc} and find that, although results are mixed, in 5 of the 8 considered models, some form of positional encoding improves the validation error over \textsc{None}. In our experiments, we therefore elect to use \textsc{Symmetric} positional encoding for all equivariant models to preserve roto-reflection equivariance and \textsc{Cartesian} positional encoding for \textsc{FNO} baselines (\textsc{FNO}, \textsc{FNO+$p4$}, and \textsc{FNO+$p4m$}). The exception is the \textsc{SWE} experiments, where the non-rectangular domain breaks the roto-reflection invariance of the \textsc{Symmetric} encoding. We therefore use \textsc{Cartesian} encoding for all models on this dataset.
        
    \subsection{Steerable parameterization of $G$-FNO}\label{app:steer}

    Steerable CNNs~\cite{cohen2017steerable} construct equivariant convolution layers using a steerable basis $\phi_1,\phi_2,\ldots,\phi_n$. The steerable basis for each layer is constructed offline by solving a linear system of equations dependent on the group representation of the input function and the desired representation of the output function. The steerable kernel $\psi$ is then given by
    \begin{equation}\label{eq:steer_def}
        \psi = \sum_{j=1}^n\alpha_j\phi_j
    \end{equation}
    where $\alpha_1,\alpha_2,\ldots,\alpha_n$ are the learnable basis coefficients. Steerable convolutions are strictly more general than group convolutions, since choosing the input and output representation as the {\em regular} representation gives an alternative method for parameterizing group convolutions~\cite{cohen2017steerable}. Furthermore, steerable convolutions can achieve equivariance to infinite groups such as continuous rotations \cite{weiler2019general}.

    To parameterize the steerable kernel $\psi$ in the frequency domain, we use the linearity of the transform as 
    \begin{equation}
        \mathcal F\psi=\sum_{j=1}^n\alpha_j\mathcal F\phi_j. 
    \end{equation}
    Thus, to learn a steerable kernel in the frequency domain, it is sufficient to learn the basis coefficients for the transform of the basis functions. This has the added benefit that all of the learnable parameters in the model, i.e., the basis coefficients, are real-valued. 
    
    In~\cref{tab:steer}, we consider a steerable parameterization of $G$-FNO for the groups $p4$ and $p4m$. These initial results suggest this parameterization is not ideal, as the models given by the original parameterization are significantly better. Future work should investigate a more effective steerable parameterization so that equivariant frequency domain convolutions can be extended to continuous groups.
    
    \section{Proofs and Background}

\subsection{The Groups $p4$ and $p4m$}\label{app:group_intro}

In this section, we characterize the groups $p4$ and $p4m$. $p4$ is the group generated by translations and $90^\circ$ rotations, while $p4m$ is $p4$ with reflections.

For us, an element $g\in p4$ is parameterized by $s_g\in \{0,1,2,3\}$ for a planar rotation and $x_g\in \R^2$ for a translation by $x_g$. We then let  $R_{s_g} = \begin{pmatrix} 0&-1\\1&0\end{pmatrix}^{s_g}$, giving a rotation  by $s_g\cdot90^\circ$. The element $g$ acts on a function $\nu:\R^2\to\R$ by applying a rotation followed by a translation as
\begin{equation}
(L_g\nu)[x] = \nu\left(R^{-1}_{s_g}x-x_g\right).
\end{equation} 
For discretized functions on a numerical grid $\delta\cdot \Z^2$, the translations $x_g$ are restricted to grid elements. For the sake of simplicity, we focus on this setting in the following, which corresponds to our discretization. The continuous case is perfectly analogous.

Next, for a function $\rho:p4\to\mathbb R$ defined on $p4$, denote $\rho(h)=\rho(s_h, x_h)$. Here, it may be helpful to picture $p4$ as $\{0,1,2,3\}\times \Z^2$, that is, a ``stack" of 4 planes. The translation coordinate $x_h$ indexes each of these planes, while the rotation coordinate $s_h$ tracks the relative pose of the corresponding plane as it pertains to rotations, with $s_h=0$ indicating that the plane is in its canonical orientation, $s_h=1$ indicating that the plane is rotated by $90^\circ$, etc. Then, $g$ transforms $\rho$ by applying a roto-translation to each of the planes and periodically incrementing the rotation coordinate $s_h$ as 
\begin{equation}
(L_g\rho)[h] = \rho\left((s_h + s_g)\bmod 4, R^{-1}_{s_g}x_h-x_g\right).
\end{equation} 
For example, for the ``slice" of $\rho$ representing the plane in the second position, i.e., rotated by $180^\circ$, rotating $\rho$ by $180^\circ$ will bring that plane to its canonical orientation, that is $(2 + 2)\bmod 4 = 0$. 

Further, for the transformation $m$ in $p4m$, $m$ is parameterized by $s_m$ and $x_m$, with $${s_m=\left(s_m^{(1)},s_m^{(2)}\right)\in\{-1,1\}\times\{0,1,2,3\}}.$$ Let $R_{s_m}$ be a 2D orthogonal matrix corresponding to a rotation by $s_m^{(2)}\cdot90^\circ$ followed by a horizontal reflection if, and only if, $s_m^{(1)} = -1$. That is, $R_{s_m} = \begin{pmatrix} s^{(1)}_m&0\\0&1\end{pmatrix}\begin{pmatrix} 0&-1\\1&0\end{pmatrix}^{s^{(2)}_m}$. Then, $m$ transforms the function $\nu$ similar to before by applying a roto-reflection followed by a translation as
\begin{equation}
(L_m\nu)[x] = \nu\left(R^{-1}_{s_m}x-x_m\right).
\end{equation} 
Lastly, for the function $\eta:p4m\to\mathbb R$ defined on $p4m$, again denote $\eta(p)=\eta(s_p, x_p)$. Similar to the $p4$ case, $x_p$ indexes into the ``stack" of eight planes comprising $p4m$, while $s_p^{(1)}$ tracks the relative pose of the corresponding plane with respect to horizontal reflections and $s_p^{(2)}$ tracks the rotational pose. Then, $m$ transforms $\eta$ as
\begin{equation}
(L_m\eta)[p] = \eta\left(s_p^{(1)}\cdot s_m^{(1)},(s_p^{(2)} + s_m^{(2)})\bmod 4, R^{-1}_{s_m}x_p-x_m\right).
\end{equation} 
For example, for the ``slice" of $\eta$ representing the plane in the pose $s_p=(-1,3)$, i.e., reflected and rotated by $270^\circ$, applying a roto-reflection by $90^\circ$ will bring that plane to its canonical orientation, that is $\left(-1\cdot -1,(3 + 1)\bmod 4\right) = (1,0)$. Note that vertical reflections are accomplished by composing a rotation and horizontal reflection.

    \subsection{Proof of Symmetry of Fourier transform to $O(n)$}\label{app:four_sym}

        \begin{lemma}
            Let $A\in \R^{n\times n}$ be an invertible matrix, $f:\R^n\to\R$ Lebesgue-integrable and $b\in\R^n$. Consider the function $f_{A,b}:\R^n\to\R$ given by $f_{A,b}(x) = f(Ax+b)$. Then
            \[
            \mathcal F(f_{(A,b)})(\xi) = \frac{e^{-2\pi i \,\langle A^{-T}\xi, b\rangle}}{|\det A|}\,\mathcal F(f)(A^{-T}\xi)
            \]
        \end{lemma}

        In particular, if $A$ is an orthogonal matrix, then $|\det A| = 1$ and $A^{-T} = A$, so for all $O\in O(n)$:
        \[
        \mathcal F(f_{(O,b)})(\xi) =  e^{-2\pi i \,\langle O\xi, b\rangle}\,\mathcal F(f)(O\xi)
        \]

        \begin{proof}
        We will use the multi-dimensional change of variables formula with the substitution $z = Ax+b$, as well as the identity $\langle \xi, Az\rangle = \langle A^T\xi, z\rangle$.
        \begin{align*}
        \mathcal F(f_{(A,b)})(\xi) &= \frac1{(2\pi)^{n/2}} \int_{\R^n} e^{-2\pi i\, \langle\xi, x\rangle} f_{(A,b)}(x)\,dx\\
            &=\frac1{(2\pi)^{n/2}\,|\det A|} \int_{\R^n} e^{-2\pi i\,\langle\xi, A^{-1}(Ax+b)\rangle + 2\pi i\,\langle \xi, A^{-1}b\rangle} f(Ax+b)\,|\det A|dx\\
            &= e^{2\pi i \langle \xi, A^{-1}b\rangle}\,\frac1{(2\pi)^{n/2}\,|\det A|} \int_{\R^n} e^{-2\pi i\,\langle\xi, A^{-1}z\rangle} f(z)\,dz\\
            &= \frac{e^{2\pi i \,\langle \xi, A^{-1}b\rangle}}{|\det A|} \,\frac1{(2\pi)^{n/2}}\int_{\R^n} e^{-2\pi i\,\langle A^{-T}\xi, z\rangle}\,f(z)\,dz\\
            &= \frac{e^{2\pi i \,\langle A^{-T} \xi, b\rangle}}{|\det A|}\,\mathcal F(f)(A^{-T}\xi).
        \end{align*}
        \end{proof}

    \subsection{Proof of Navier-Stokes Closure to Action of $O(2)$}\label{app:ns_close}

                    \label{lem:rt_sol}                    
        
        \begin{lemma}
            \label{lem:ns_close_app}
            Let $U\subseteq\R^2$ be a domain or $U = \T^2 = \R^2/\Z^2$ the flat torus/periodic square in which we identify the top/bottom and left/right sides. 
            
            Suppose that the functions $u: U\to\R^2$, $w, f: U\to\R$ solve the partial differential equation
    
            \begin{align}
                \label{eqn:ns_app}
                \begin{split}
                    \partial_t w(x,t) + u(x,t)\cdot\nabla w(x,t)& =\nu\Delta w(x,t) + f(x)
                    \\
                    \nabla\cdot u(x,t) & = 0
                \end{split}
            \end{align}  
            
            Take $R$ to be an orthogonal matrix describing a rotation or reflection. Then the functions $u_R : R^{-1}U\to\R^2$ and $w_R, f_R:R^{-1}U\to\R$ also satisfy~\cref{eqn:ns_app}, where:
    
            \begin{align}
                u_R(x, t) & \coloneqq R^\top u(Rx, t)
                \\
                w_R(x, t) & \coloneqq w(Rx, t)
                \\
                f_R(x) & \coloneqq f(Rx)
            \end{align}
            
        \end{lemma}
        
        Note that if $f$ is invariant to the action of $O(2)$, then $f_R=f$, and thus, this result implies that $u_R$ and $w_R$ solve~\cref{eqn:ns_app}. 
        
        \cref{lem:ns_close_app} is most meaningful in our context if the domain $U$ is invariant under the rotation $R$, i.e.\ if $R^{-1}U = U$. This is true if $U=\R^2$, $U = B_1(0)$ is a disk in $\R^2$, or if $U=\R^2$ and $R$ is a reflection or a rotation by $0^\circ, 90^\circ, 180^\circ$ or $270^\circ$. The case $U = \T^2$ is associated with the study of problems that are periodic in the coordinate directions.
    
        \begin{proof}
            For brevity of notation, we omit the time variable $t$. 
            
            To show that $u_R, w_R$ and $f_R$ satisfy~\cref{eqn:ns_app}, it is enough to show that:
    
            \begin{align}
                (u_R\cdot \nabla w_R)(x) & = (u\cdot \nabla w)(Rx) \label{eqn:u_dot_nabla_w}
                \\
                (\Delta w_R)(x) & = (\Delta w)(Rx) \label{eqn:laplacian}
                \\
                (\nabla \cdot u_R)(x) & = (\nabla\cdot u)(Rx) \label{eqn:div}
            \end{align}
    
            First, to show~\cref{eqn:u_dot_nabla_w}, let $D(\nu)\in\R^{d\times q}$ denote the Jacobian matrix for $\nu:\R^q\to\R^d$. Then:
    
            \begin{align*}
                (u_R \cdot \nabla w_R)(x) 
                & = u(Rx)^\top R \left(D\left(w(Rx)\right)R\right)^\top
                \\
                & = (u\cdot\nabla w)(Rx)
            \end{align*}
    
            Next, to show~\cref{eqn:laplacian}, let $\delta_{j,k}$ be the Kronecker delta, which is $1$ if $j=k$ and $0$ otherwise:
    
            \begin{align*}
                \Delta w_R(x) 
                & = \sum_{i=1}^2\frac{\partial}{\partial x_i}\left(\frac{\partial}{\partial x_i}w(Rx)\right)
                \\
                & = \sum_{i=1}^2\sum_{j=1}^2\frac{\partial}{\partial x_i}\left(\frac{\partial w}{\partial x_j}(Rx)\sum_{k=1}^2\frac {\partial(R_{j,k}x_k)}{\partial x_i}\right)
                \\
                 & = \sum_{i=1}^2\sum_{j=1}^2\frac{\partial}{\partial x_i}\left(\frac{\partial w}{\partial x_j}(Rx)R_{j,i}\right)
                 \\            
                & = \sum_{i=1}^2\sum_{j=1}^2\sum_{k=1}^2R_{j,i}\frac{\partial w}{\partial x_k\partial x_j}(Rx)\sum_{l=1}^2\frac{\partial (R_{k,l}x_l)}{\partial x_i}
                \\
                & = \sum_{i=1}^2\sum_{j=1}^2\sum_{k=1}^2R_{j,i}\frac{\partial w}{\partial x_k\partial x_j}(Rx)R_{k,i}
                \\
                & = \sum_{j=1}^2\sum_{k=1}^2\frac{\partial w}{\partial x_k\partial x_j}(Rx)(RR^\top)_{j,k}
                \\
                & = \sum_{j=1}^2\sum_{k=1}^2\frac{\partial w}{\partial x_k\partial x_j}(Rx)\delta_{j,k}
                \\
                & = (\Delta w)(Rx)
            \end{align*}
    
            Lastly, to show~\cref{eqn:div}:
    
            \begin{align*}
                \nabla\cdot u_R(x) 
                & = \sum_{i=1}^2 \frac\partial{\partial x_i}\left(R^\top u(Rx)\right)_i
                \\
                & = \sum_{i=1}^2 \sum_{j=1}^2\frac\partial{\partial x_i}R_{j,i} u_j(Rx)
                \\
                & = \sum_{i=1}^2 \sum_{j=1}^2\sum_{k=1}^2 R_{j,i}\frac{\partial u_j}{\partial x_k}(Rx)\sum_{l=1}^2\frac{\partial (R_{k,l}x_l)}{\partial x_i}
                \\
                & = \sum_{i=1}^2 \sum_{j=1}^2\sum_{k=1}^2 R_{j,i}\frac{\partial u_j}{\partial x_k}(Rx) R_{k,i}
                \\
                & = \sum_{j=1}^2\sum_{k=1}^2 \frac{\partial u_j}{\partial x_k}(Rx) (R R^\top)_{j,k}
                \\
                & = \sum_{j=1}^2\sum_{k=1}^2 \frac{\partial u_j}{\partial x_k}(Rx) \delta_{j,k}
                \\
                & = (\nabla\cdot u)(Rx)
            \end{align*}
            
        \end{proof}

    \section{Rollout Visualizations}\label{sec:roll_vis}

\begin{figure}[h]
  \centering
    \hspace*{\fill}%
  \subfloat[Vorticity]{\includegraphics[width=0.45\textwidth]{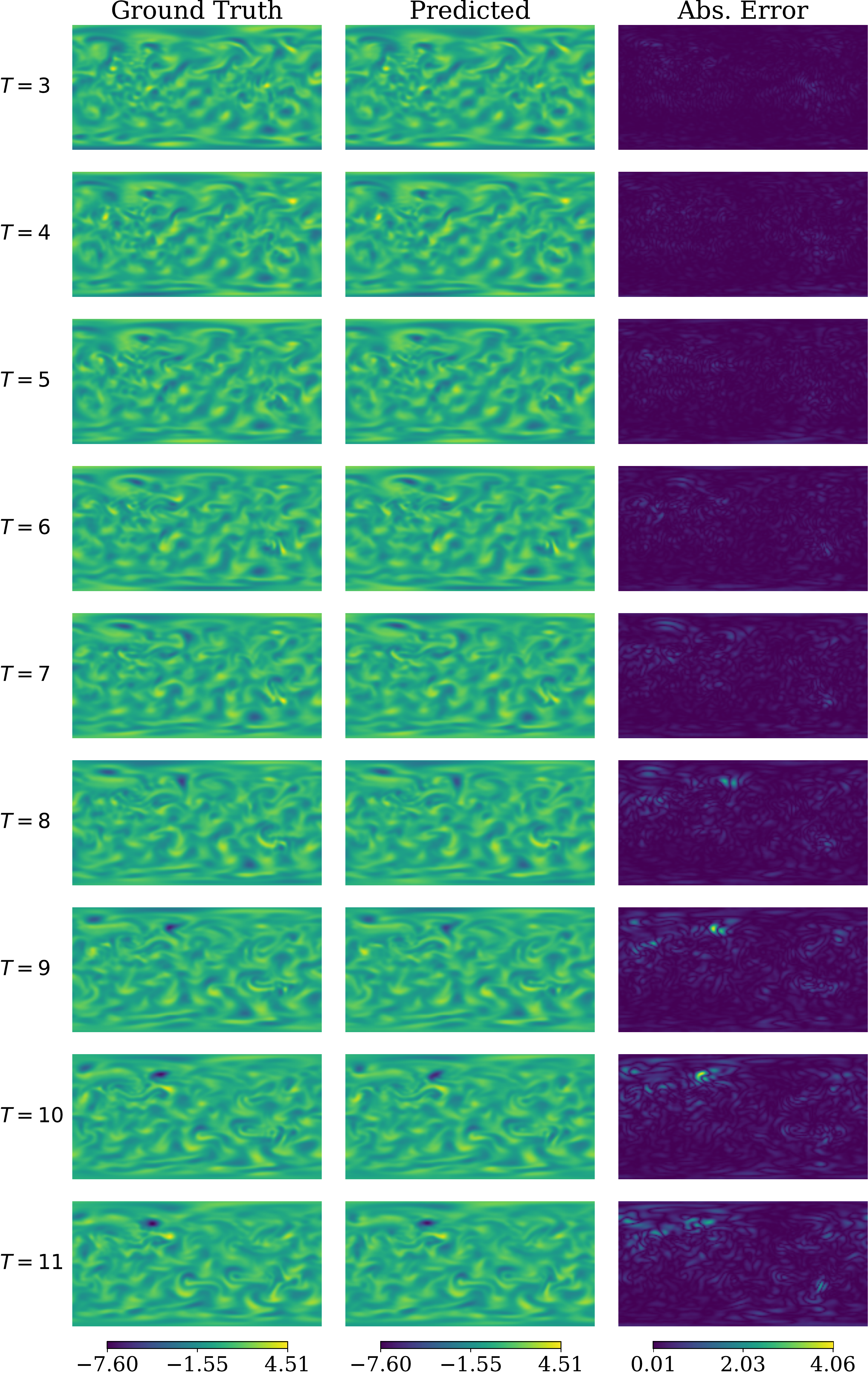}}
  \hfill%
  \subfloat[Pressure]{\includegraphics[width=0.45\textwidth]{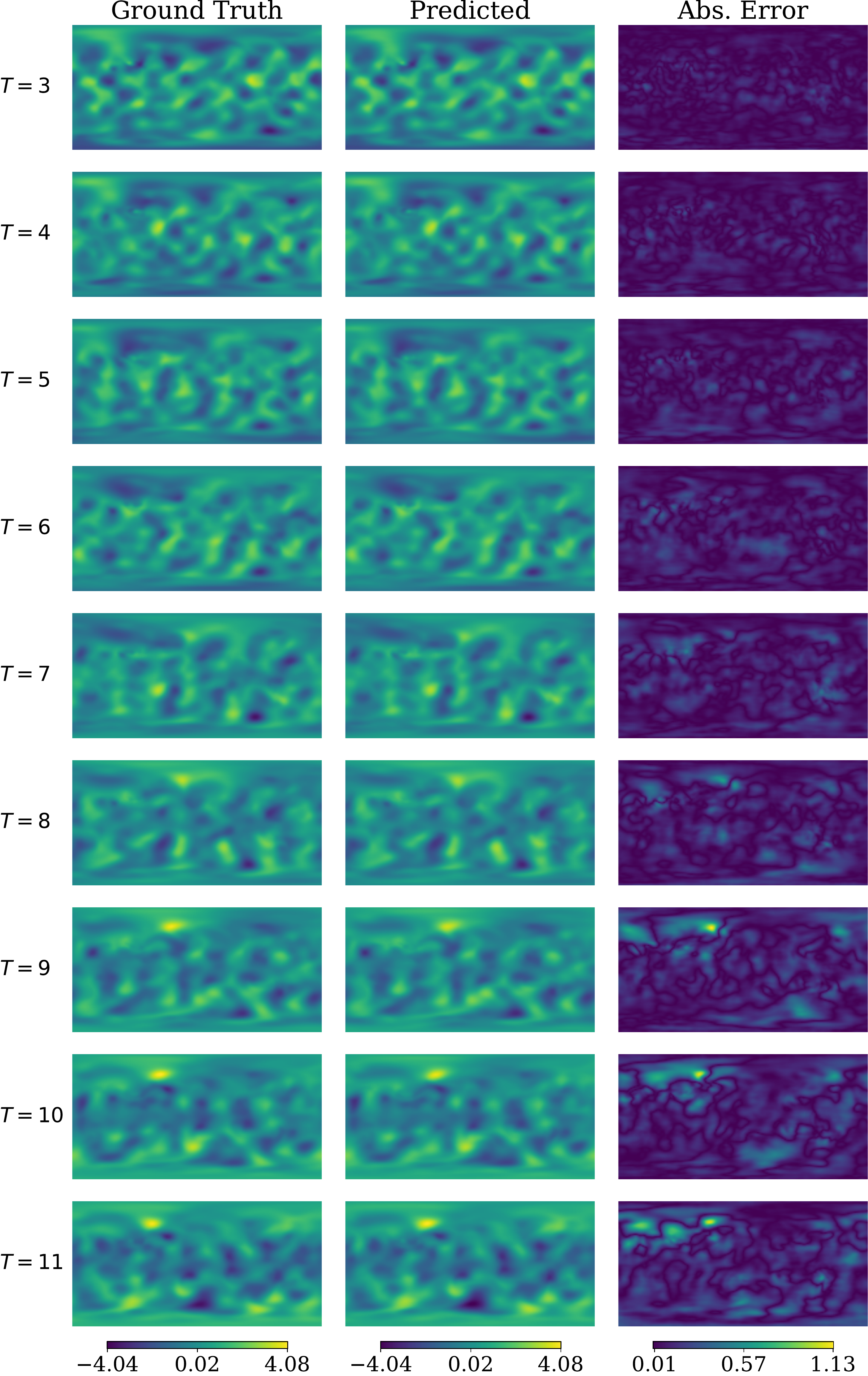}}
  \hspace*{\fill}%
  \caption{Rollout of \textsc{SWE}.}
  \label{fig:rollout_SWE_arena}
\end{figure}

\begin{figure}[h]
  \centering
    \hspace*{\fill}%
  \subfloat{\includegraphics[width=0.34\textwidth]{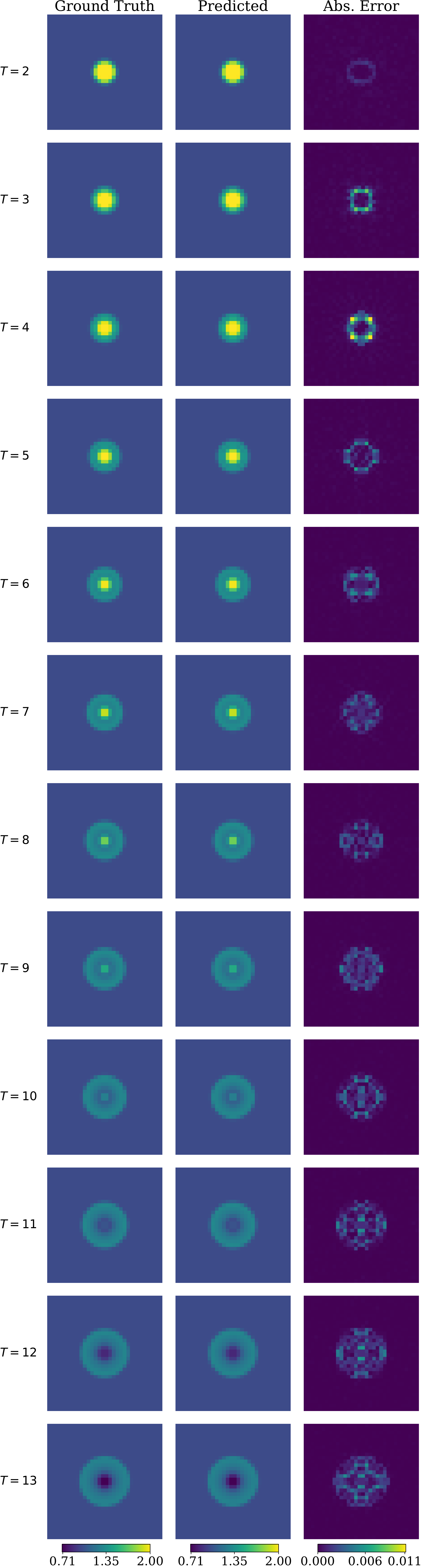}}
  \hfill%
  \subfloat{\includegraphics[width=0.34\textwidth]{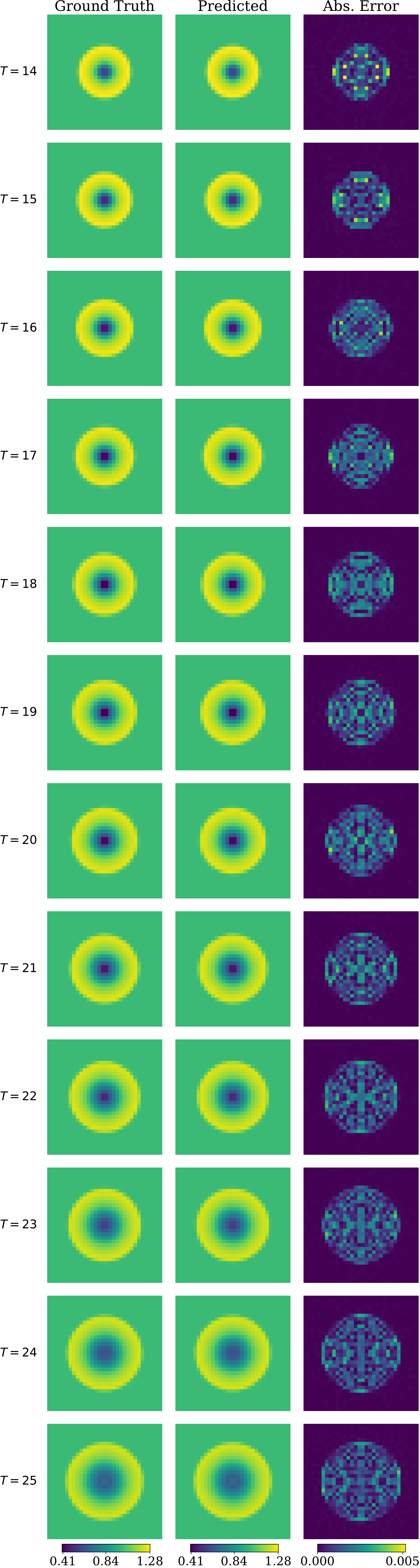}}
  \hspace*{\fill}%
  \vspace{-5pt}
  \caption{Rollout of \textsc{SWE-sym}. Note that the scales of the error bars on the left and the right differ.}
  \label{fig:rollout_SWE_bench}
\end{figure}

\begin{figure}[h]
  \centering
    \hspace*{\fill}%
  \subfloat{\includegraphics[width=0.4\textwidth]{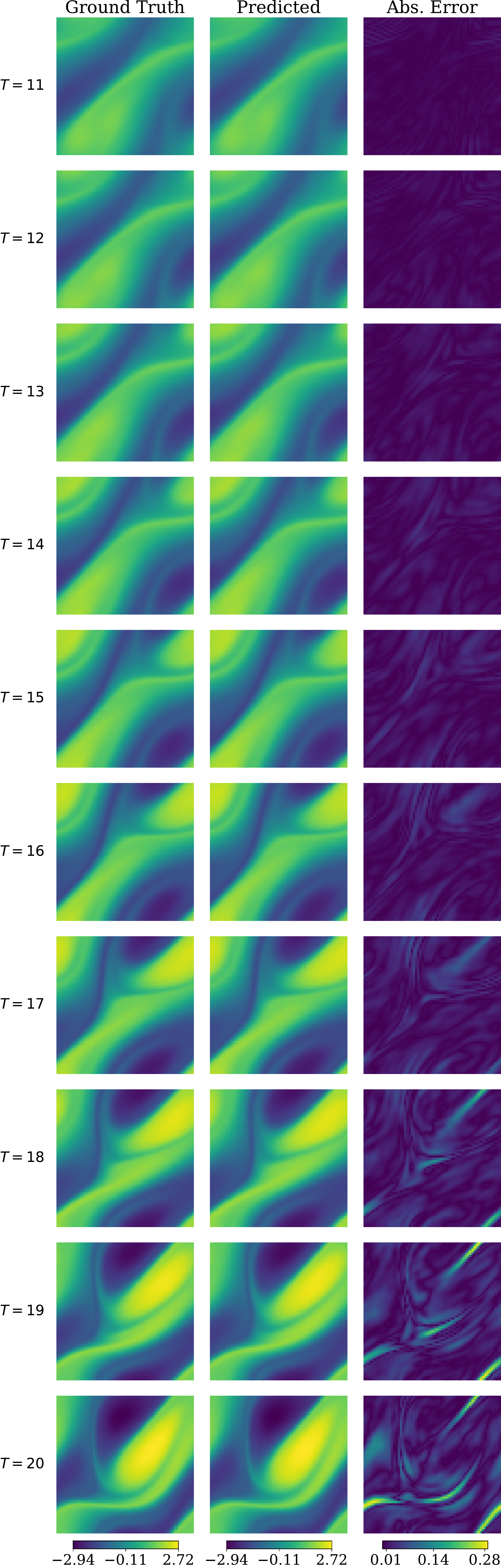}}
  \hfill%
  \subfloat{\includegraphics[width=0.4\textwidth]{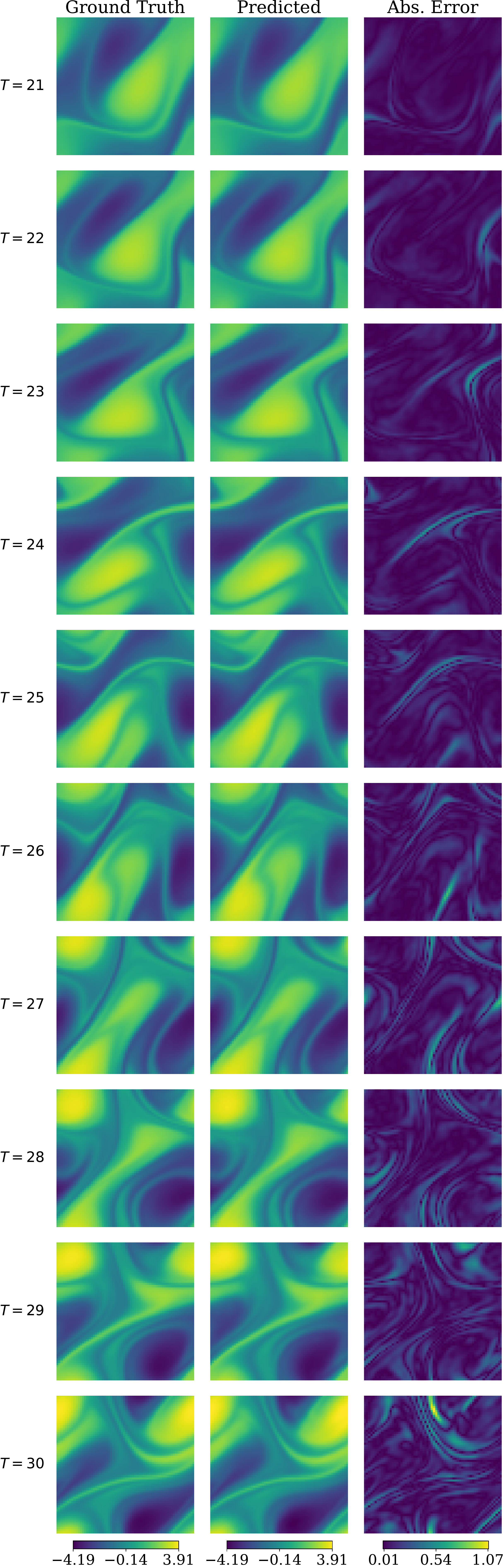}}
  \hspace*{\fill}%
  \caption{Rollout of Navier-Stokes with non-symmetric forcing (\textsc{NS}). Note that the scales of the error bars on the left and the right differ.}
  \label{fig:rollout_ns_nonsym}
\end{figure}

\begin{figure}[h]
  \centering
    \includegraphics[width=0.45\textwidth]{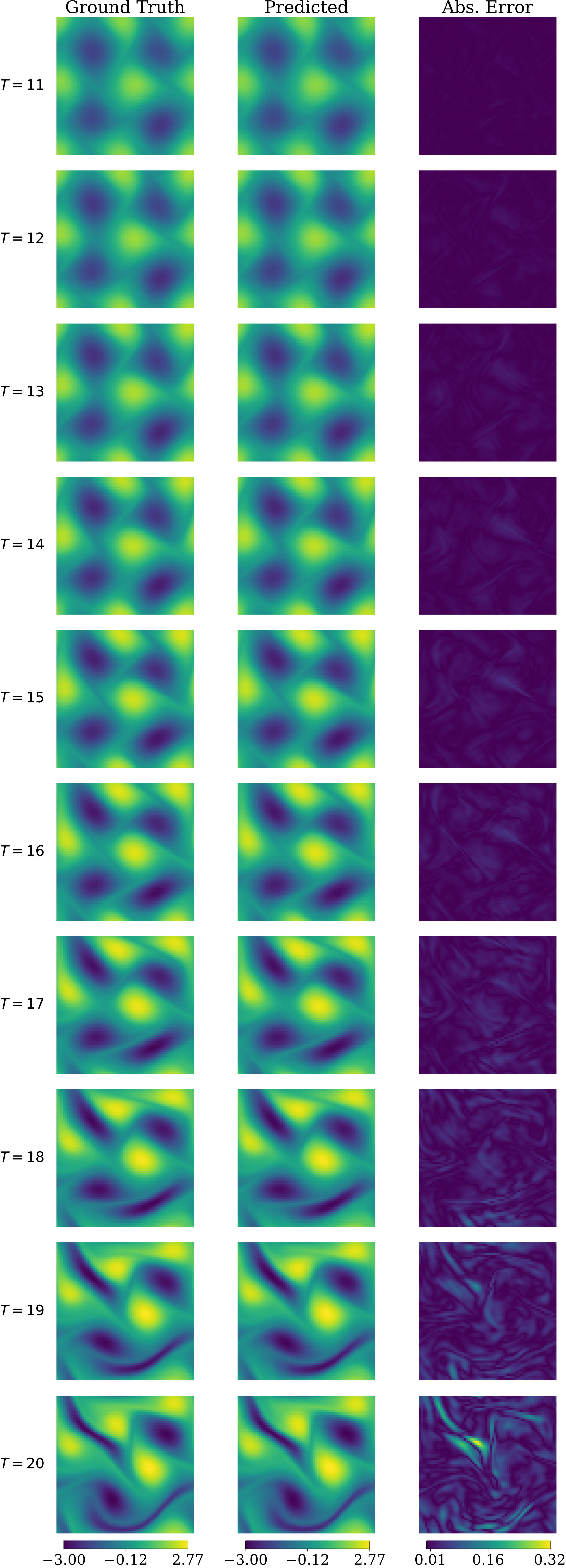}
    \caption{Rollout of Navier-Stokes with symmetric forcing (\textsc{NS-Sym}).}
  \label{fig:rollout_ns_sym}
\end{figure}

    In this section, we randomly select a trajectory from the test set and visualize the ground truth rollout alongside the rollout predicted by the \textsc{2D} version of \textsc{$G$-FNO-$p4$} trained to make predictions autoregressively. We visualize \textsc{SWE} in~\cref{fig:rollout_SWE_arena}, \textsc{SWE-sym} in~\cref{fig:rollout_SWE_bench}, \textsc{NS} in~\cref{fig:rollout_ns_nonsym}, and \textsc{NS-sym} in~\cref{fig:rollout_ns_sym}.
    
\end{document}